\documentclass[final,1p,times]{elsarticle}

\usepackage{epsfig}
\usepackage{amssymb}
\usepackage{lineno}
\usepackage{multirow}
\usepackage{subfigure}
\usepackage{enumerate}
\usepackage[table,xcdraw]{xcolor}
\usepackage{amsmath}
\usepackage{epstopdf}
\usepackage{enumitem,url}
\usepackage{fixltx2e}

\begin{document}

\pagenumbering{arabic}
\baselineskip 13.4pt
\begin{frontmatter}

\title{A Deep Convolutional Neural Network for Background Subtraction}

\author[a1]{M. Babaee\corref{cor1}}
\author[a1]{D. Dinh}
\author[a1]{G. Rigoll}

\address[a1]{Institute for Human-Machine Communication, Technical Univ. of Munich, Germany}

\begin{abstract}
In this work, we present a novel background subtraction system that uses a deep Convolutional Neural Network (CNN) to perform the segmentation. With this approach, feature engineering and parameter tuning become unnecessary since the network parameters can be learned from data by training a single CNN that can handle various video scenes. Additionally, we propose a new approach to estimate background model from video. For the training of the CNN, we employed randomly $5\%$ video frames and their ground truth segmentations taken from the Change Detection challenge 2014(CDnet 2014). We also utilized spatial-median filtering as the post-processing of the network outputs. Our method is evaluated with different data-sets, and the network outperforms the existing algorithms with respect to the average ranking over different evaluation metrics. Furthermore, due to the network architecture, our CNN is capable of real time processing.

\end{abstract}

\end{frontmatter}
\section{Introduction}
\label{sec:introduction}
With the tremendous amount of available video data, it is important to maintain the efficiency of video based applications to process only relevant information. Most video files contain redundant information such as background scenery, which costs a huge amount of storage and computing resources. Hence, it is necessary to extract the meaningful information, e.g. vehicles or pedestrians, to deploy those resources more efficiently.
Background subtraction is a binary classification task that assigns each pixel in a video sequence with a label, for either belonging to the background or foreground scene~\cite{varadarajan2015regionmog,st2015subsense,barnich2011vibe}.\\
Background subtraction, which is also called change detection, is applied to many advanced video applications as a pre-processing step to remove redundant data, for instance in tracking or automated video surveillance~\cite{sajid2015colorspacessinglegaussian}. In addition, for real-time applications, like tracking, the algorithm should be capable of processing the video frames in real-time.\\
One simple example of the application of a background subtraction method is the pixel-wise subtraction of a video frame from its corresponding background image. After being compared with the difference threshold, pixels with a larger difference than a certain threshold value are labeled as foreground pixels, otherwise as background pixels. Unfortunately, this strategy will yield poor segmentation due to the dynamic nature of the background, that is induced by noise or illumination changes. For example, due to lighting changes, it is common that even pixels belonging to the background scene can have intensities very different from their other pixels in the background image and they will be falsely classified as foreground pixels as a consequence. Thus, sophisticated background subtraction algorithms that assure robust background subtraction under various conditions must be employed.\\
In the following sections, the difficulties in this area, our proposed solution for background subtraction and our contributions will be illustrated.\\

\subsection{Challenges}
\label{sec:challenges}
The main difficulties that complicate the background subtraction process are:\\
\\
\textbf{Illumination changes:} When scene lighting changes gradually (e.g. moving clouds in the sky) or instantly (e.g. when the light in a room is switched on), the background model usually has a illumination different from the current video frame and therefore yields false classification.\\
\\
\textbf{Dynamic background:} The background scene is rarely static due to movement in the background (e.g. waves, swaying tree leaves), especially in outdoor scenes. As a consequence, parts of the background in the video frame do not overlap with the corresponding parts in the background image, hence, the pixel-wise correspondence between image and background is no longer existent.\\
\\
\textbf{Camera jitter:} In some cases, instead of being static, it is possible that the camera itself is frequently in movement due to physical influence. Similar to the dynamic background case, the pixel locations between the video and background frame do not overlap anymore. The difference in this case is that it also applies to non-moving background regions.\\
\\
\textbf{Camouflage:} Most background subtraction algorithms work with pixel or color intensities. When foreground objects and background scene have a similar color, the foreground objects are more likely to be (falsely) labeled as background.\\
\\
\textbf{Night Videos:} As most pixels have a similar color in a night scene, recognition of foreground objects and their contours is difficult, especially when color information is the only feature in use for segmentation.\\
\\
\textbf{Ghosts/intermittent object motion:} Foreground objects that are embedded into the background scene and start moving after background initialization are the so-called \textit{ghosts}. The exposed scene, that was covered by the ghost, should be considered as background. In contrast, foreground objects that stop moving for a certain amount of time, fall into the category of intermittent object motion. Whether the object should be labeled as foreground or background is strongly application dependent. As an example, in the automated video surveillance case, abandoned objects should be labeled as foreground.\\
\\
\textbf{Hard shadows:} Dark, moving shadows that do not fall under the illumination change category should not be labeled as foreground.\\

In this work, we follow the trend of Deep Learning and apply its concepts to background subtraction by proposing a CNN to perform this task. We justify this approach with the fact that background subtraction can be performed without temporal information, given a sufficiently good background image. With such a background image, the task itself breaks down into a comparison of a image-background pair. Hence, the input samples can be independent among each other, enabling a CNN to perform the task with only spatial information. The CNN is responsible for extracting the relevant features from a given image-background pair and performs segmentation by feeding the extracted features into a classifier. In order to get more spatially consistent segmentation, post-processing of the network output is done by spatial-median filtering and or a fully connected CRF framework. Due to the use of a CNN, no parameter tuning or descriptor engineering is needed. 

To train the network, a large amount of labeled data is required. Fortunately, due to the process of background subtraction, by comparing an image with its background, it is not necessary to use images of a full scene for training. It is also possible to train the network via subsets of a scene, i.e. with patches of image-background pairs, since the procedure also holds for image patches. As a consequence, we can extract enough training samples from a limited amount of labeled data.

To the best of our knowledge, background subtraction algorithms that use CNN are scene specific to this day, i.e. a CNN can only perform satisfying background subtraction on a single scene (that was trained with scene specific data) and also lacks the ability to perform the segmentation in real time. Our proposed approach yields a universal network that can handle various scenes without having to retrain it every time the scene changes. As a consequence, one can train the network with data from multiple scenes and hence increase the amount of training data for a network. Also, by using the proposed network architecture, it is possible to process video frames in real-time with conventional computing resources. Therefore, our approach can be considered for real-time applications. 

The outline of this paper is as follows: In Section~\ref{sec:related_work}, early and recent algorithms for background subtraction are presented. In Section~\ref{sec:approach}, we explain our proposed approach for background subtraction. Here, we first describe our proposed approach to estimate background image and next we illustrate our CNN for background subtraction. In Sections~\ref{sec:experiments}, we describe the experimental evaluation of the algorithm including the used datasets, the evaluation metrics and the obtained results followed by detailed discussion and analysis. Finally, in Section~\ref{sec:conclusion}, we conclude our work and provide future work with some ideas.

\section{Backround and Related work}
\label{sec:related_work}

\begin{figure}[t]
	\centering
	\includegraphics[width=0.85\linewidth]{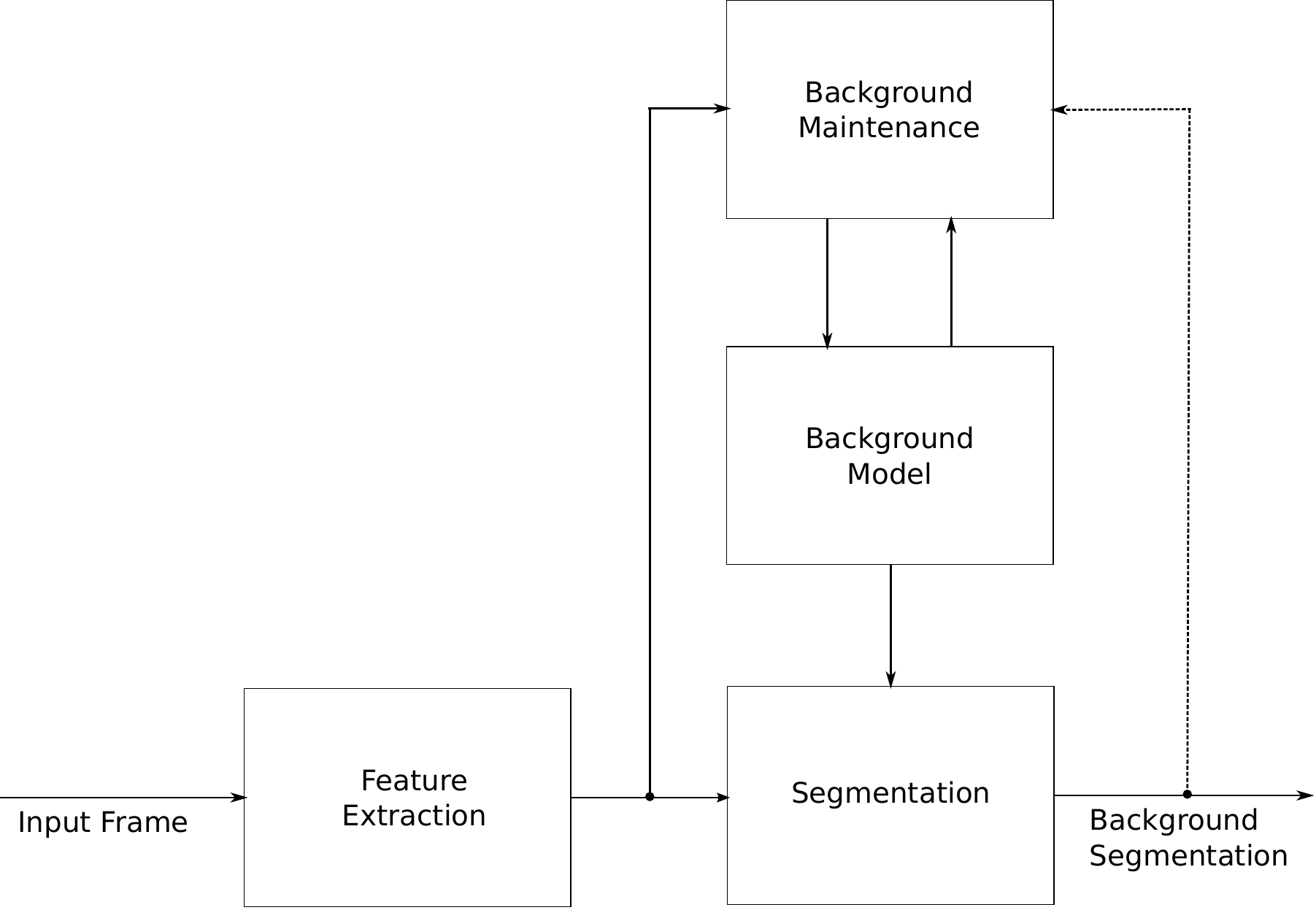}
    \caption[Block diagram of basic background subtraction algorithm]{Block diagram of basic background subtraction algorithms. The dashed line is optional and not existent in some algorithms}
    \label{fig:background_subtraction_block_diagram}
\end{figure}

The majority of background subtraction algorithms are composed of several processing modules which are explained in the following sections (see Figure~\ref{fig:background_subtraction_block_diagram}).\\
\\
\textbf{Background Model:} The background model is essential for the background subtraction algorithm. In general, the background model is used as a reference to compare with the incoming video frames. Furthermore, the initialization of the background model plays an important role since  video sequences are normally not completely free of foreground objects during the bootstrapping phase. As a consequence, the model gets corrupted by including foreground objects into the background model, which leads to false classifications.\\
\\
\textbf{Background Model Maintenance:} In reality, background is never completely static but changes over time. There are many strategies to adapt the background model to these changes by using the latest video frames and/or previous segmentation results. Trade-offs must be found in the adaption rate, which regulates how fast the background model is updated. High adaption rate leads to noisy segmentation due to the sensitivity to small or temporary changes. Slow adaption rate, however, yields an outdated background model and therefore false segmentation. Selective update adapts the background model with pixels that were classified as background. In that case, deadlock situations can occur by not incorporating misclassified pixels into the background model, i.e. once a background pixel is falsely classified as foreground, it would never be used to update the background and would always be considered as a foreground pixel. On the other hand, by using all pixels as in the blind update strategy, such deadlock situations can be prevented but will also distort the background model since foreground pixels are incorporated into the background model.\\
\\
\textbf{Feature extraction:}
In order to compare the background image with the video frames, adequate features that represent relevant information must be selected. Most algorithms use gray scale or RGB intensities as features. In some cases, pixel intensities along with other hand engineered features (e.g. \cite{heikkila2006lbp} or \cite{bilodeau2013lbsp}) are combined.
Also, the choice of the feature region is important. One can extract the features over pixels, blocks or patterns. Pixel-wise features often yield noisy segmentation results since they do not encode local correlation, while block-wise or pattern-wise features tend to be insensitive to slight changes.\\
\\
\textbf{Segmentation:} With the help of a background model, the respective video frames can be processed. Background segmentation is performed by extracting the features from corresponding pixels or pixel regions of both frames and using a distance measure, e.g. the Euclidean distance, to measure the similarity between those pixels. After being compared with the similarity threshold, each pixel is either labeled as background or foreground.\\
\\
The combination of those building blocks forms an overall background subtraction system. The robustness of the system is always dependent and limited by the performance of each individual block, i.e. it can not be expected to perform well if one module delivers poor performance.

\begin{figure}[!t]
\centering
\includegraphics[width=1.1\linewidth]{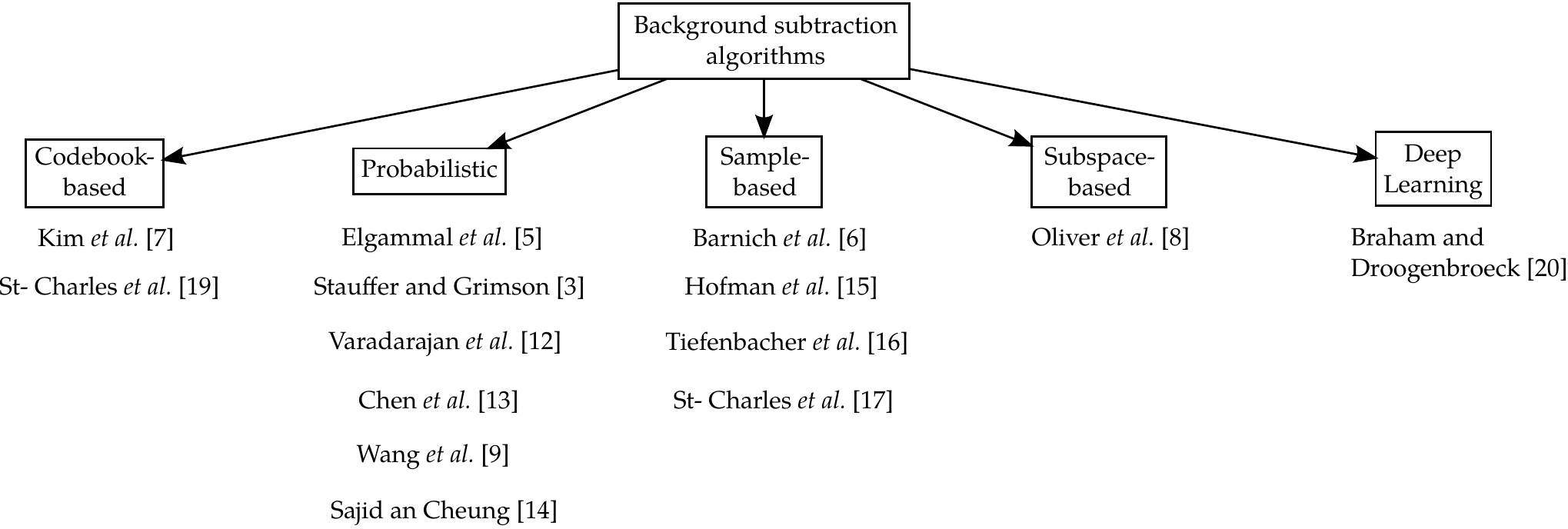}
\caption[Overview of background subtraction algorithms]{Categorization of the outlined background subtraction algorithms}
\label{fig:bs_algorithm_categorization}
\end{figure}

Background subtraction is a well studied field, therefore there exists a vast number of algorithms for this purpose (see Figure. \ref{fig:bs_algorithm_categorization}). Since most of the top performing methods at present are based on the early proposed algorithms, some of which are outlined in the beginning. Subsequently a few of the current methods for background subtraction will be introduced.

\subsection{Early Approaches}
\label{subsec:early_approaches}
Stauffer and Grimson \cite{stauffer1999mog} proposed a method that models the background scene with Mixture of Gaussian (MoG), also called Gaussian Mixture Model (GMM). It is assumed that each pixel in the background is drawn from a Probability Distribution Function (PDF) which is modeled by a GMM, also pixels are assumed to be independent from their neighboring pixels. Incoming pixels from video frames are labeled as background if there exists a Gaussian in the GMM, where the distance between its mean and the pixel lies within a certain bound. For learning the parameters, that maximize the likelihood, the authors proposed an online method that approximates the Expectation Maximization (EM) algorithm \cite{dempster1977em_algorithm}.\\
Elgammal \textit{et al.} \cite{elgammal2000kde} introduced a probabilistic non-parametric method to model the background. Again it is assumed that each background pixel is drawn from a PDF. The PDF for each pixel is estimated with Kernel Density Estimation (KDE).\\
In contrast, Barnich \textit{et al.} \cite{barnich2011vibe} propose a sample based method for background modeling. The background model consists of pixel samples from past video frames. For each pixel location, the number of past pixel samples is stored as \mbox{N} (e.g. $\mbox{N=20}$). To perform the background segmentation at each pixel position, the incoming pixel is compared with the $\mbox{N}$ pixel samples from the background model. If there exist at least $\mbox{K}$ samples in the model with a distance to the incoming pixel smaller than a global threshold $\mbox{R}$, the pixel is classified as background pixel, otherwise as foreground pixel.\\
Another approach is introduced by Kim \textit{et al.} \cite{kim2005codebook}, where the background model for each pixel position is modeled by a codebook. During bootstrapping phase, incoming pixels build up codewords that consist of intensity, color and temporal features. After initialization, those codewords form a codebook for subsequent segmentation. The intensity and color of incoming pixels are compared with those of the codewords in the codebook. After calculating the distances between the pixels and the codewords and comparing them with the threshold value , either a foreground label (if no match was found) or a background label is assigned. In the latter case, the matching codeword is updated with the respective background pixel.\\
Oliver \textit{et al.} \cite{oliver2000pcasubspace} use subspace learning to compress the background, the so called eigenbackground. From $\mbox{N}$ images, the mean and the covariance matrix are calculated. After a PCA of the covariance matrix, a projection matrix is set up with $\mbox{M}$ eigenvectors, corresponding to the $\mbox{M}$ largest eigenvalues. Incoming images are compared with their projection onto the eigenvectors. After calculating the distances between the image and the projection and comparing them with the corresponding threshold value, foreground labels are assigned to pixels with large distances.

\subsection{Recent Approaches}
\label{subsec:recent_approaches}
Based on \cite{stauffer1999mog}, Wang \textit{et al.} \cite{wang2014fluxtensormog} use GMM to model the background scene, in addition, single Gaussians are employed for foreground modeling. By computing flux tensors \cite{bunyak2007fluxtensor}, which depict variations of optical flow within a local 3D spatio-temporal volume, blob motion is detected. With the combination of the different information from blob motion, foreground models and background models, moving and static foreground objects can be spotted. Also, by applying edge matching \cite{evangelio2011edge_matching}, static foreground objects can be classified as ghosts or intermittent motions.\\
Varadarajan \textit{et al.} \cite{varadarajan2015regionmog} introduce region-based MoG for background subtraction to tackle the sensitivity to dynamic background . MoGs are used to model square regions within the frame, instead of pixel-wise feature modeling, the features are extracted from sub-blocks within the square regions.\\
With the same purpose to tackle dynamic background movement, Chen \textit{et al.} \cite{chen2015sharable} propose a method that applies MoG in a local region. Models for foreground and background are learned with pixel-wise MoG. To label a given pixel, a $N\times N$ region around the pixel is searched for a MoG that shows the highest probability for the center pixel. If a foreground model depicts the highest probability, the center pixel is labeled as foreground and vice versa.\\
To cope with sudden illumination changes, Sajid and Cheung \cite{sajid2015colorspacessinglegaussian} create multiple background models with single Gaussians and employ different color representations. Input images are pixel-wise clustered into $K$ groups with the K-means algorithm. For each group and each pixel location a single Gaussian model is built by calculating the mean and variance for a cluster. For incoming pixels, the matching models are selected by taking the models among the $K$ models that show the highest normalized cross-correlation to the pixels. The segmentation is done for each color channel for RGB and YCbCr color representations, which yields 6 segmentation masks. By combining all available segmentation masks the background segmentation is then performed.\\
Hofmann \textit{et al.} \cite{hofmann2012pbas} developed an algorithm that improved the method from Barnich \textit{et al.} \cite{barnich2011vibe} by replacing the global threshold value $\mbox{R}$ with an adaptive threshold  $\mbox{R(x)}$, which is dependent on the pixel location and a metric of the background model. The metric was named ''background dynamics'' by the authors. With the additional information from the background dynamics, the thresholds $\mbox{R(x)}$ and the model update rate are adapted by a feedback loop. High variance in the background yields a larger threshold value and vice versa, therefore it can cope with dynamic background and highly structured scenes. Tiefenbacher \textit{et al.} \cite{tiefenbacher2014pbas_pid} improved this method by controlling the updates of the pixel-wise thresholds with a PID controller instead of a fixed value that was used before.\\
St-Charles \textit{et al.} \cite{st2015subsense} also improved the method by using Local Binary Similarity Patterns (LBSP)~\cite{bilodeau2013lbsp} as additional features to pixel intensities and slight changes in the update heuristic of the thresholds and the background model.\\
They also proposed another method \cite{st2015pawcs}, which is  similar \cite{st2015subsense}, but works with codewords, or background words as they call it in their publication. Pixel intensities, LBSP features and temporal features are combined to a background word. To perform segmentation, incoming pixels' LBSP features together with their intensities are compared with the corresponding background words. After a two stage comparison, 1), color and structure threshold, 2), temporal feature threshold, the pixel is either labeled as foreground or background depending on the threshold results. All thresholds are dynamically updated in a feedback loop using a background dynamics measure.\\
A novel approach for background subtraction with the use of CNN was proposed by Braham and Droogenbroeck \cite{braham2016deep}. They used a fixed background model, which was generated from a temporal median operation over $N$ video frames. Afterwards, a scene-specific CNN is trained with corresponding image patches from the background image, video frames and ground truth pixels, or alternatively, segmentation results coming from other background subtraction algorithms. After extracting a patch around a pixel, feeding the patch through the network and comparing it with the score threshold, the pixel is assigned with either a background or a foreground label. However, due to the over-fitting that is caused by using highly redundant data for training, the network is scene-specific, i.e. can only process a certain scenery, and needs to be retrained for other video scenes (with scene-specific data).\\

\section{Approach}
\label{sec:approach}
\begin{figure}[!t]
\centering
\includegraphics[width=\linewidth, height=175pt]{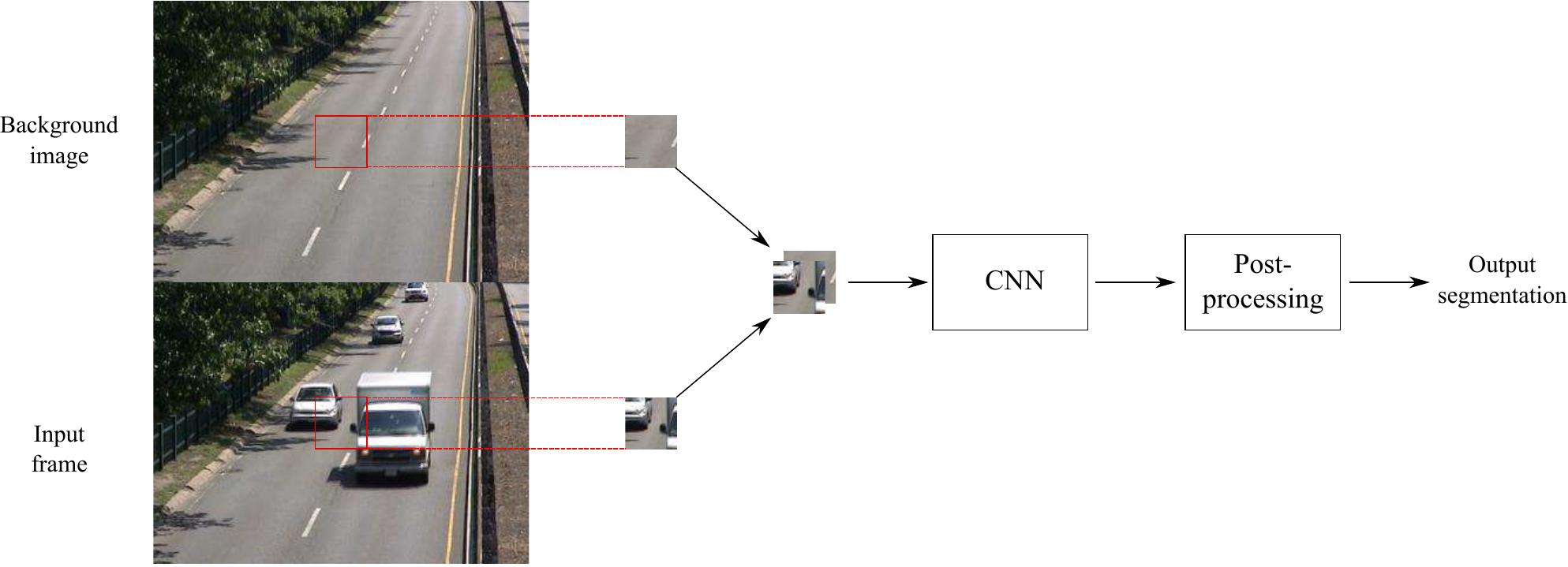}
\caption[Background subtraction system overview]{Background subtraction system overview: The input frame along with the corresponding background image are patch-wise processed. After merging the individual patches into a single output frame, the output frame is post-processed yielding the final output segmentation.}
\label{fig:background_subtraction_system_overview}
\end{figure}

In this section, we introduce our proposed method that consists of 1) a novel algorithm for background model (image) generation; 2) a novel CNN for background subtraction and 3) post-processing of the networks output using median filter. The complete system is illustrated in Figure \ref{fig:background_subtraction_system_overview}. We use a background image to perform background subtraction from the incoming frames. Matching pairs of image patches from the background image and the video frames are extracted and fed into a CNN. After reassembling the patches into the complete output frame, it is post-processed, yielding the final segmentation of the respective video frame.\\
In the following sections, we introduce our algorithm to get the background images from the video frames. Furthermore, we illustrate our CNN architecture and the training procedure. At last, we discuss our employed post-processing strategies to improve the network output.

\begin{figure}
\centering
\includegraphics[width=0.85\linewidth]{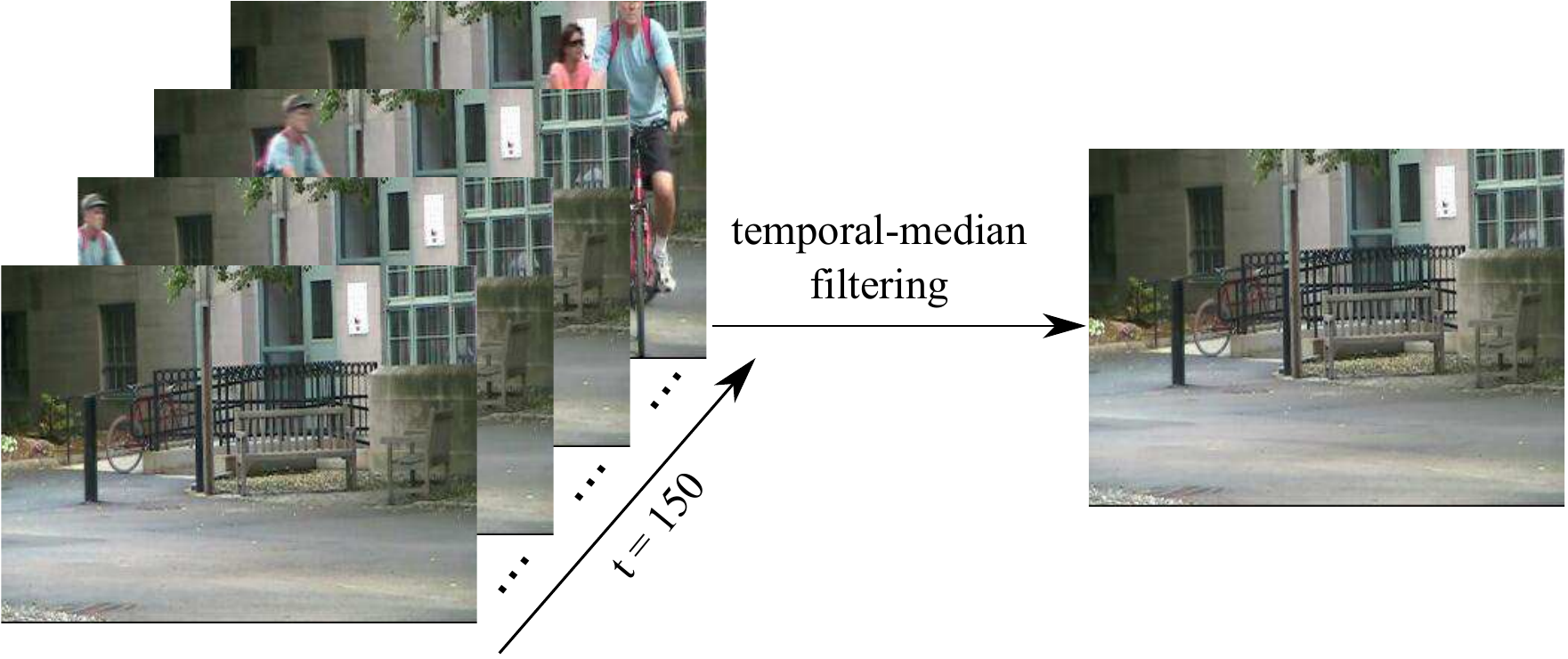}
\caption[Temporal-median filtering of a video sequence]{Temporal-median filtering over 150 frames of a video sequence. Since each background pixel is visible for at least 75 frames, the temporal-median filtering yields the background image without foreground objects.}
\label{fig:temporal_median}
\end{figure}

\subsection{Background Image Generation}
\begin{figure}[htb]
	\centering
	\includegraphics[width=0.85\linewidth]{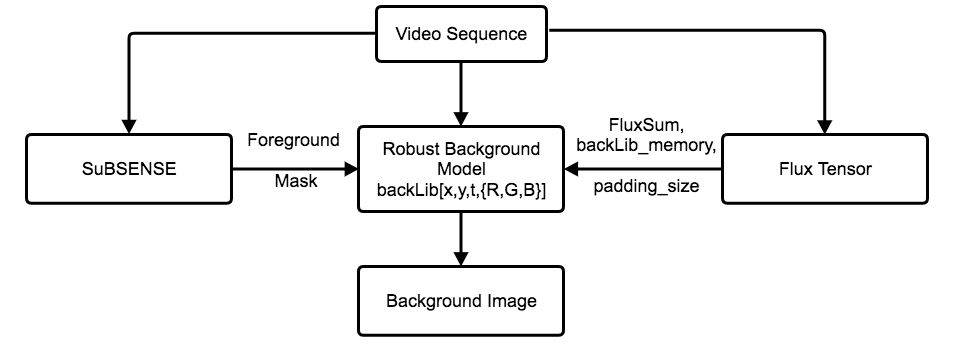}
	\caption{Block diagram of Robust Background Model Algorithm}
	\label{fig:Block}
\end{figure}

We propose a new approach to generate background model which is illustrated in Figure.~\ref{fig:Block}. Here we combine the segmentation mask from SuBSENSE algorithm~\cite{st2015subsense} and the output of Flux Tensor algorithm~\cite{wang2014fluxtensormog}, which can dynamically change the parameters used in the background model based on the motion changes in the video frames. The block diagram of the robust background model algorithm is given in Figure.~\ref{fig:Block}. The details of each block is explained in the following sections.   
\subsubsection{Foreground Segmentation Using SuBSENSE}
Perhaps, the simplest way to get background image from a video sequence is to perform pixel-wise temporal median filtering. However, using this method, the background model will be quickly corrupted if there are a lot of moving objects in some video frames, and hence, the pixel values of foreground object will eventually negatively affect the quality of the background model. This requires us to distinguish the foreground pixels and background pixels, and for the background model we only use the background pixel values to perform temporal median filter. To this end, we use the SuBSENSE algorithm. This method relies on spatio-temporal binary features as well as color information to detect changes. This allows camouflaged foreground objects to be detected more easily while most illumination variations are ignored. Besides, instead of using manually-set, frame-wide constants to dictate model sensitivity and adaptation speed, it uses pixel-level feedback loops to dynamically adjust the method's internal parameters without user intervention. The adjustments are based on the continuous monitoring of model fidelity and local segmentation noise levels.
	%
The output of SuBSENSE algorithm will be a binary image which contains the classification information of the current video frame. The foreground objects are marked as white pixel, and black pixels represent the pixels belonging to background model.
\begin{figure}[htb]
	\centering
	\begin{tabular}{cc}
	\includegraphics[width=0.4\linewidth]{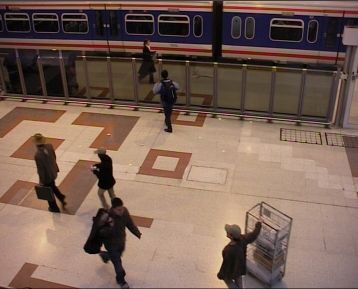} &
	\includegraphics[width=0.4\linewidth]{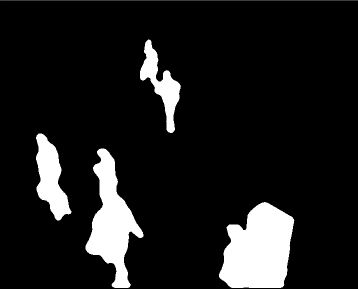}\\
	(a) & (b)
	\end{tabular}

	\caption{(a) Video input frame; (b) segmentation output of SuBSENSE}
	
	\label{fig:Sub_Out}
\end{figure}

\subsubsection{Background Pixel Library}
\begin{figure}[htb]
	\centering
	\includegraphics[width=\columnwidth]{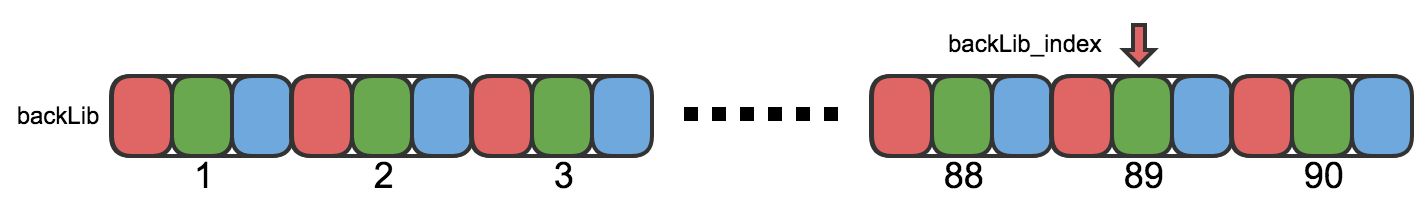}
	\caption{Background pixel library for one pixel location}
	\label{fig:BackLib}
\end{figure}

\noindent
Based on the foreground mask image from SuBSENSE, we build a background pixel library,$BL$, for each pixel location in the frame. The idea is that we only store the pixel values from the current frame in the background pixel library, when they are classified by SuBSENSE algorithm as background pixels. An illustration of background pixel library is shown in Figure.~\ref{fig:BackLib}. 
Here, we only keep the last 90 background pixel values from the video sequences. After the library is complete, the oldest background pixel value is replaced with the newest one. For this purpose, we have an indicator, $bi$, which is a pointer that points to the oldest background pixel value in the library. Each time, if a pixel value is classified by SuBSENSE as background, then it will be stored in the background library at the location where $bi$ points to, and then we move $bi$ to the next position in the library. To generate the background image, we calculate the average value over a certain memory length of the pixel values in the background pixel library. The memory length is defined using the variable $bm$. The background pixel at location $(x,y)$ and at color channel $k$ is defined by $BK(x,y,k)$, whose value is calculated using following equation
\begin{equation}
BK(x,y,k)=\frac{\sum\limits_{i=bi-bm}^{bi} BL(x,y,i,k)}{bm},~~k=\{R,G,B \}.\label{eq:avg_bak}
\end{equation}
But if we use a fixed memory length $bm$ over the entire video sequence, we will get either blurry or outdated background model if the camera is moving or if there are some intermittent objects in the video sequence. These two cases are illustrated in Figure.~\ref{fig:FixedLength}. As we can see in the following figure, the first row shows the scenario that the camera is constantly rotating, the calculated background image is shown on the right hand side. In this case, the background image becomes very blurry if we use a fixed average length $bm$, or in the second row, the car on the left side stops at the same location for 80 percent of the frames in the video sequence, so the car will naturally be classified as background by SuBSENSE. Then if the car starts to move, the calculated background image using fixed memory length will not be updated because the pixel values of the car are still stored in the background library.
\begin{figure}[htb]
	\centering
	\begin{tabular}{cccc}
	\includegraphics[width=0.22\linewidth]{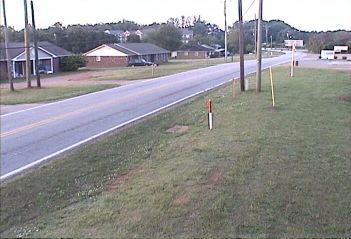} &
	\includegraphics[width=0.22\linewidth]{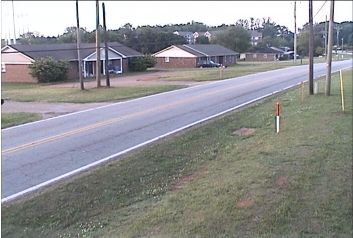} &
	\includegraphics[width=0.22\linewidth]{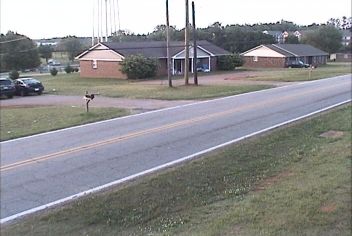} &
	\includegraphics[width=0.22\linewidth]{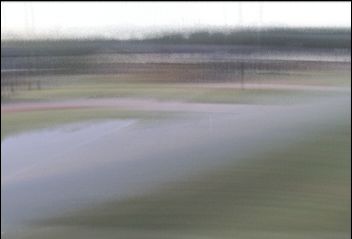} \\
	(a) & (a') & (a'') & (a''') \\	
	\includegraphics[width=0.22\linewidth]{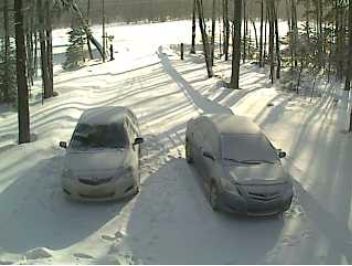} &
	\includegraphics[width=0.22\linewidth]{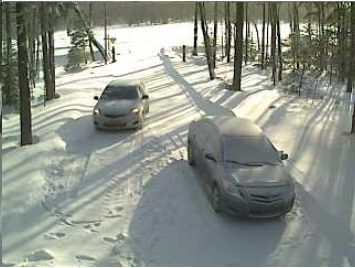} &
	\includegraphics[width=0.22\linewidth]{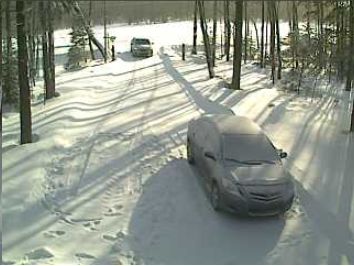} &
	\includegraphics[width=0.22\linewidth]{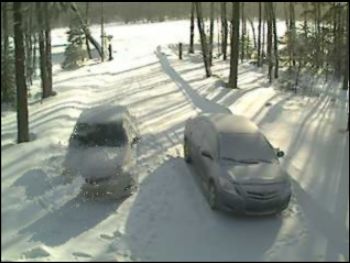} \\

  (b) & (b') & (b'') & (b''') \\	
	
	\end{tabular}	
	
	\caption{Bad Background Image Model When Using Fixed Average Length. On the first row, the first 3 images ($a-a''$) are the video input frames (Nr. 80,120,160) from continuousPan sequence of CDnet data-set. The first 3 images on the second row ($b-b''$) are the video input frames(Nr. 1800,1950,2100) from winterDriveaway sequence. The left images on the first ($a'''$) and on the second row ($b'''$) are the default background models from SuBSENSE with fixed memory length.}
	\label{fig:FixedLength}
\end{figure}\newline

\subsubsection{Motion Detection Using Flux Tensor}
\begin{figure}[htb]
	\centering
	\begin{tabular}{cc}
	\includegraphics[width=0.44\linewidth]{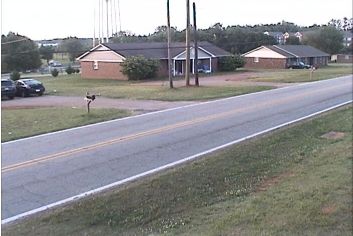} &
	\includegraphics[width=0.44\linewidth]{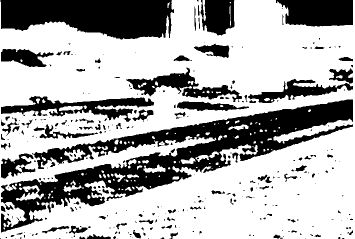}\\
	(a) & (b)\\
	\end{tabular}
	\caption{(a) input video; (b) the output of Flux Tensor}
	\label{fig:FluxTensor}
\end{figure}

In order to have adaptive memory length based on the motion of the camera and objects in the video frames, we need to have a motion detector. A commonly used motion detector is the method using flux tensor. Compared with standard motion detection algorithms, the advantage of using flux tensor is that the motion information can be directly computed without expensive eigenvalue decompositions. Flux tensor represents the temporal variation of the optical flow field within the local 3D spatio-temporal volume \cite{wang2014fluxtensormog}, where in the expanded matrix form, flux tensor is written as
\begin{equation}
J_F=\begin{bmatrix}
	\int_{\Omega}\{\frac{\partial^2 I}{\partial x \partial t}\}^2 dy  & \int_{\Omega} \frac{\partial^2 I}{\partial x \partial t} \frac{\partial^2 I}{\partial y \partial t}dy  & 
	\int_{\Omega} \frac{\partial^2 I}{\partial x \partial t} \frac{\partial^2 I}{\partial t^2 }dy  \\
	\int_{\Omega} \frac{\partial^2 I}{\partial y \partial t} \frac{\partial^2 I}{\partial x \partial t}dy & 
	\int_{\Omega} \{\frac{\partial^2 I}{\partial y \partial t}\}^2 dy& 
	\int_{\Omega} \frac{\partial^2 I}{\partial y \partial t} \frac{\partial^2 I}{\partial t^2}dy  \\
	\int_{\Omega} \frac{\partial^2 I}{\partial t^2} \frac{\partial^2 I}{\partial x \partial t}dy & 
	\int_{\Omega} \frac{\partial^2 I}{\partial t^2} \frac{\partial^2 I}{\partial y \partial t}dy & \int_{\Omega} \{ \frac{\partial^2 I}{\partial t^2} \}^2dy  
\end{bmatrix}. \label{eq:flux}
\end{equation}
The elements of the flux tensor incorporate information about temporal gradient changes which leads to efficient discrimination between stationary and moving image features. The trace of the flux tensor matrix which can be compactly written and computed as
\begin{equation}
trace(J_F)=\int_{\Omega} \lVert \frac{\partial}{\partial t} \bigtriangledown I \rVert^2 dy, \label{eq:trace_flux}
\end{equation}
can be directly used to classify moving and non-moving regions without eigenvalue decompositions. In our approach, the output of Flux Tensor algorithm is a binary image, which contains the motion information of the video frames. An example is shown in Figure.~\ref{fig:FluxTensor}. The white pixel in the binary image indicates that the pixel at this location is moving either temporally or spatially. 
\newline
\noindent
Next, we define a new variable called $F_s$ as:
\begin{equation}
F_s = \frac{N_w}{W \times H}, \label{eq:fluxsum}
\end{equation}
where $N_w$ represents the number of white pixels in Flux tensor while $W$ and $H$ represent the width and height of the image respectively. The variable $F_s$ presents how many percent of the pixels in the current video frame are moving. Large $F_s$ means either the camera is moving or there is a large object in the frame which starts to move. In this case, we need to decrease the memory length $bm$. If $F_s$ is relative small, then it means the background is steady and we can use a large memory length to suppress the noise. Using $F_s$, we dynamically increase or decrease the value of $bm$. The relation between $bm$ and $F_s$ is given as follows
\begin{equation}
bm = \begin{cases}
5 & if~F_s \geq 0.25 \\
\displaystyle 5+ \frac{F_s-0.02}{0.25-0.02} \cdot 85 & if~0.02 < F_s < 0.25 \\
90 & if~F_s \leq 0.02
\end{cases}. \label{eq:fluxsum_backLib_memory}
\end{equation}
In order to avoid the noise in $F_s$ and resulting noise in $bm$, there is a low pass filter structure applied to the value of $F_s$, the low pass filter is defined as follows
\begin{equation}\begin{aligned}
F_s(t)=\alpha \cdot F_s(t) + (1-\alpha) \cdot F_s(t-1) 
 \end{aligned}\label{eq:fluxsum_lowpass}
\end{equation}
where
\begin{equation}\begin{aligned}
\alpha = \begin{cases} 0.2 &if~F_s(t)<F_s(t-1) \\ 
0.01 &if~F_s(t)>F_s(t-1)\end{cases} \end{aligned}\label{eq:fluxsum_lowpass_alpha}
\end{equation}

Here, $F_s(t)$ means the value of $F_s$ at time~$t$. Note that the different value of~$\alpha$ is based on the fact whether $F_s$ is increasing or decreasing. The reason for this is that after a dramatic decrease of $F_s$ we want to increase $F_s$ slower, in order to let $B$ be updated with new background pixel values. 
\subsubsection{Foreground Mask Padding}
\begin{figure}[htb]
	\centering
	\begin{tabular}{cc}
	\includegraphics[width=0.5\linewidth]{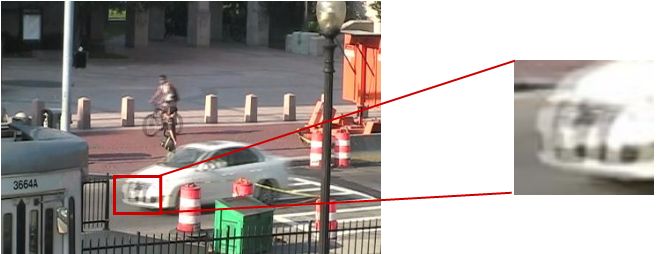} &
	\includegraphics[width=0.5\linewidth]{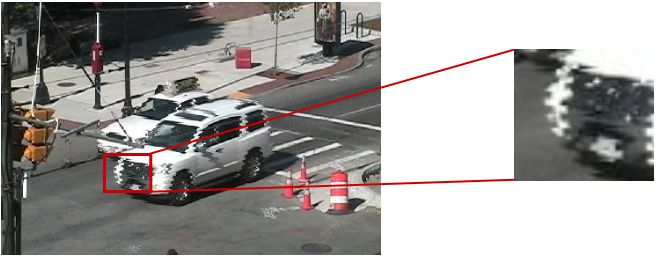}\\
	(a) & (b)\\
	\end{tabular}
	\caption{Semi-transparency and motion blur around moving objects}
	\label{fig:Motion}
\end{figure}

\begin{figure}[htb]
	\centering
	\begin{tabular}{ccc}
	\includegraphics[width=0.3\linewidth]{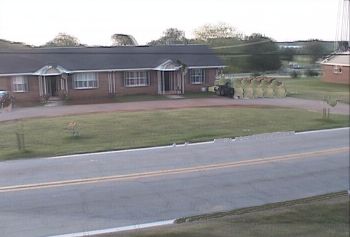}&
	\includegraphics[width=0.3\linewidth]{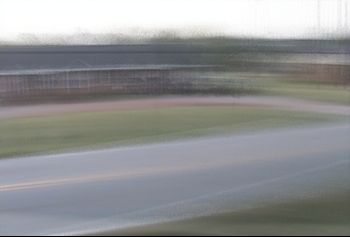}&
	\includegraphics[width=0.3\linewidth]{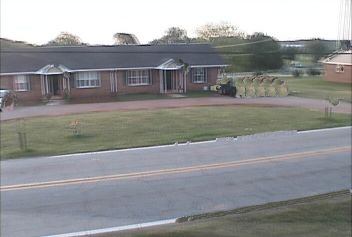}\\
		(a) & (a') & (a'')\\
	\includegraphics[width=0.3\linewidth]{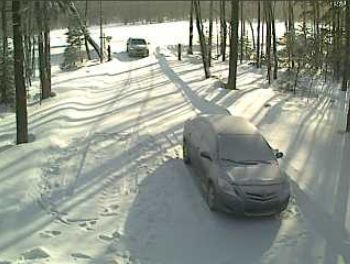}&
	\includegraphics[width=0.3\linewidth]{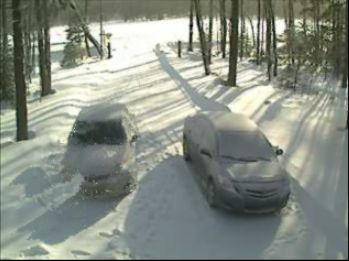}&
	\includegraphics[width=0.3\linewidth]{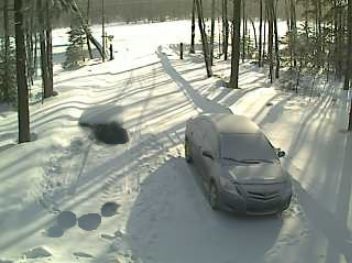}\\
	(b) & (b') & (b'')\\
	\end{tabular}

	\caption{Comparison of background models obtained by SubSENSE and the propose approach. Each row represents the results of one sample. The first column shows the input video, the second column shows the background model from SubSENSE,  and the theirs column shows the background image from proposed approach.}
	\label{fig:Comparison}
\end{figure}

Due to the low quality of some  surveillance cameras, there will be some undesired pixel values around moving objects in the form of semi-transparency and motion blur, which are illustrated in Figure.~\ref{fig:Motion}.
\noindent
In these two cases, the pixels near segmentation mask of SuBSENSE will probably be false negatives. This means that even the pixels near the segmentation mask are classified as background, but they are actually corrupted by foreground moving pixels with semi-transparency and motion blur. In this case, we should not add this background pixel to the background library. For this purpose we need to perform a padding around the foreground mask with the size defined by using variable $P_w$, which means if a pixel value is classified as background but the pixels around it within the radius of path size $P_s$ are classified as foreground, then this background pixel will be disregarded. The $P_s$ should also be dynamically adjusted using the output of flux tensor. For instance, if $F_s$ is large, then we need to increase $P_s$, because with more moving pixels the phenomenon of semi-transparency and motion blur will also increase. Figure.~\ref{fig:Comparison} shows the comparison between the background model from SuBSENSE and the robust background model obtained by the proposed approach. 

\subsection{Network Architectures and Training}
We train the proposed CNN with background images obtained by the SuBSENSE algorithm \cite{st2015subsense}. Both networks are trained with pairs of RGB image patches from video and background frames and the respective ground truth segmentation patches. Before introducing the network architecture, we outline the data preparation step for the network training. Afterwards, we will illustrate our architecture for our CNN and explain the training procedure in detail.

\label{network_architectures}

\subsubsection{Data preparation}
\begin{figure}[t]
	\centering
	\begin{tabular}{ccc}
  \includegraphics[height=180pt,width=0.3\linewidth]{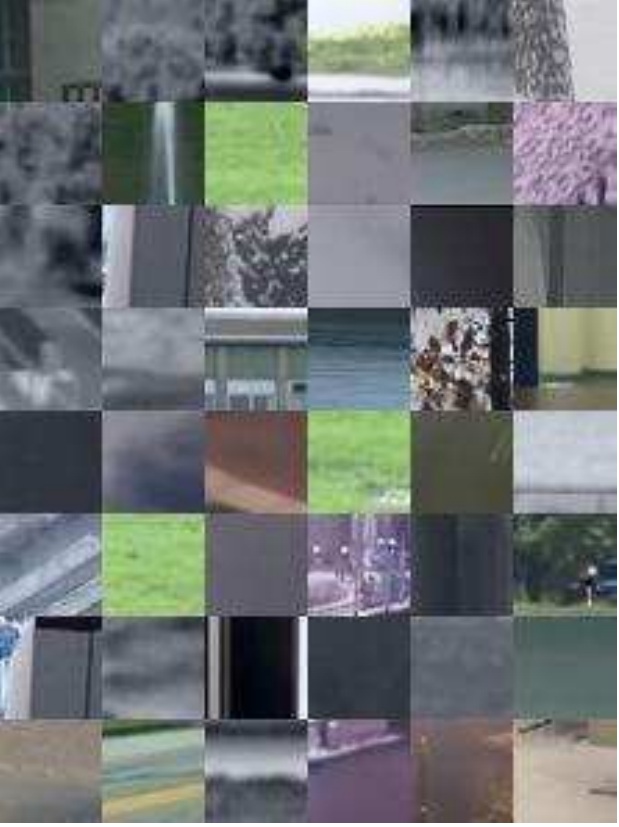} \label{fig:backgroundminibatch}&
	\includegraphics[height=180pt,width=0.3\linewidth]{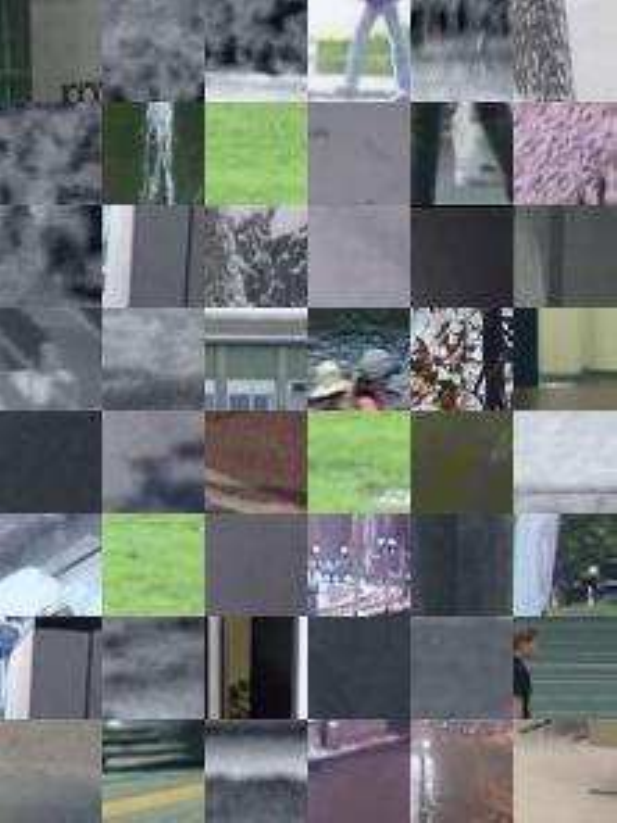}\label{fig:videominibatch}&
 	\includegraphics[height=180pt,width=0.3\linewidth]{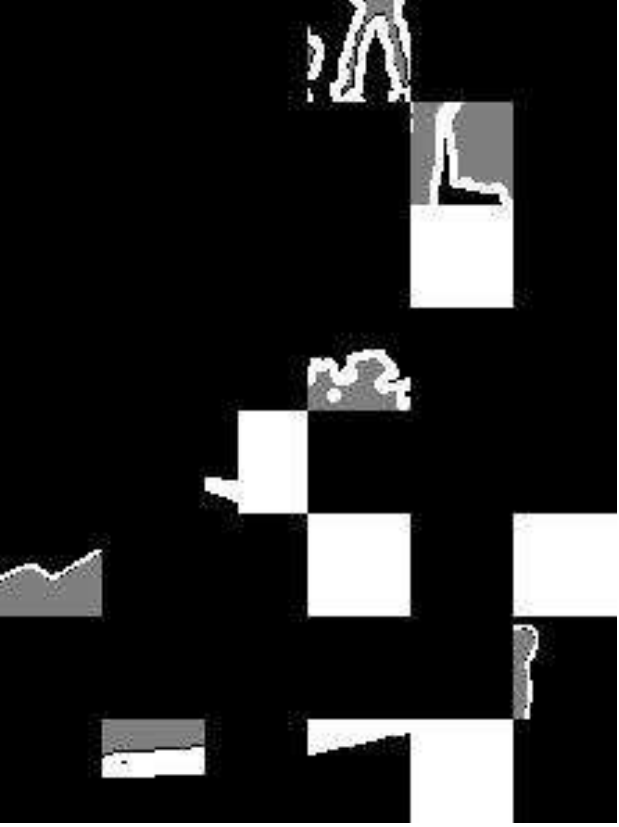}\label{fig:labelminibatch}\\
	(a)&(b)&(c)\\
	\end{tabular}
      \caption{Visualization of a mini batch of size 48: (a) Background patches, generated with temporal-median filtering. (b) Corresponding input patches from video frames. (c) Ground truth patches, black pixels are background pixels, gray pixels are foreground pixels and white pixels are not of interest, i.e. those are not incorporated into the loss.}
    \label{fig:mini_batch}
\end{figure}

For the training of the CNN, we use random data samples (around 5 percent) from the CDnet 2014 data-set \cite{wang2014cdnet}, which contains various challenging video scenes and their ground truth segmentation. We prepare one set of background images that is obtained by the proposed algorithm. Since we want the network to learn representative features, we only utilize video scenes that do not challenge the background model, i.e. the background scene should not change significantly in a video. Therefore, we exclude all videos from certain categories for training (see Table \ref{tab:cdnet2014_training_categories}).\\
We work with RGB images for background and video frames. Before we extract the patches, all employed frames are resized to the dimension $240\times320$ and the RGB intensities are rescaled to $\left[0,1\right]$. Furthermore, zero padding is applied before patch extraction to avoid boundary effects. The training data consist of triplets of matching patches from video, ground truth and background frames of size $37\times37$ that are extracted with a stride of 10 from the employed training frames. An example mini-batch of training patches is shown in Figure. \ref{fig:mini_batch}.\\
As widely recommended, we perform mean subtraction on the image patches before training, but we discard the division by the standard deviation, since we are working with RGB data and therefore each channel has the same scale.\\

\begin{table}[t]
\centering
\begin{tabular}{|c|c|}
\hline
\multicolumn{2}{|c|}{\textbf{CDnet 2014 \cite{wang2014cdnet} categories}} \\ \hline
badWeather                 & X   \\ \hline
baseline                   & X   \\ \hline
cameraJitter               & X   \\ \hline
dynamicBackground          & X   \\ \hline
intermittentObjectMotion   &     \\ \hline
lowFramerate               &     \\ \hline
nightVideos                & X   \\ \hline
PTZ                        &     \\ \hline
shadow                     & X   \\ \hline
thermal                    & X   \\ \hline
turbulence                 & X   \\ \hline
\end{tabular}
\caption[Categories of CDnet 2014]{Categories of CDnet 2014 \cite{wang2014cdnet}: Crosses indicate categories including all their video sequences that were considered for network training.}
\label{tab:cdnet2014_training_categories}
\end{table}

\subsubsection{Network Architecture and Optimization}
\begin{figure}[t]
	\centering
  \includegraphics[width=0.98\linewidth]{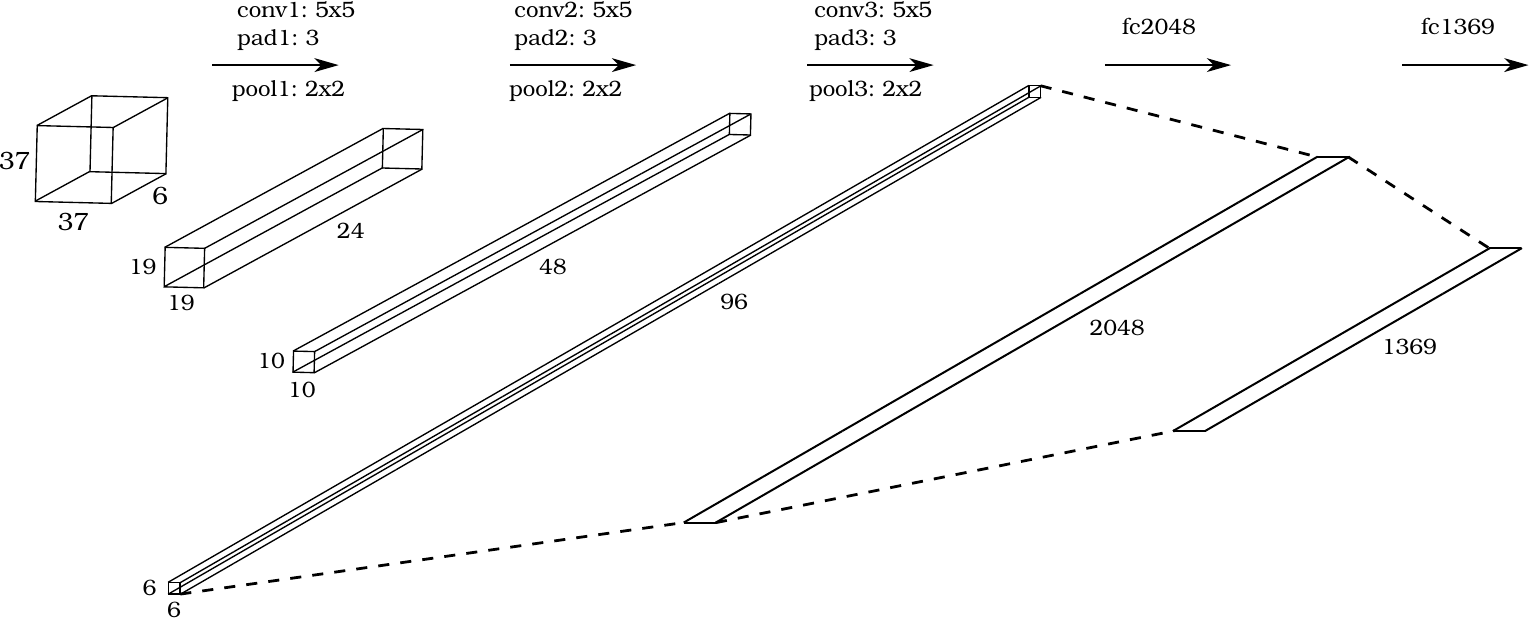}
    \caption[Illustration of proposed CNN]{Illustration of proposed CNN: The network contains 3 convolutional layers and a 2 layer MLP.}
    \label{fig:cnn_architecture}
\end{figure}

The architecture of the proposed CNN is illustrated in Figure \ref{fig:cnn_architecture}. The network contains 3 convolutional layers and a 2-layer Multi Layer Perceptron (MLP). We use the Rectified Linear Unit (ReLU) as activation function after each convolutional layer and the Sigmoid function after the last fully connected layer. In addition, we insert batch normalization layers \cite{ioffe2015batchnormalization} before each activation layer. A batch normalization layer stores the running average of the mean and standard deviation from its inputs. The stored mean is subtracted from each input of the layer and also division by the standard deviation is performed. It has been shown that by applying batch normalization layers, over-fitting is decreased and also higher learning rates for training are achieved.\\
We train the networks with mini batches of size 150 via RMSProp \cite{hinton2012lecturermsprop} and a learning rate of $\alpha=2.5* 10^{-3}$. For the loss function, we choose the Binary Cross Entropy (BCE), which is defined as follows:

\begin{equation}
BCE(x,y) =  -(x * \log{y} + (1 - x) * \log{(1 - y)}).
\end{equation}
The BCE is calculated between the network outputs and the corresponding vectorized ground truth patches of size $37*37 = 1369$. Boundaries of foreground objects and pixels that do not lie in the region of interest are marked in the ground truth segmentations. These pixels are ignored in the cost function.

\subsection{Post-Processing}
\label{model_postprocessing}
\begin{figure}
	\centering
	\begin{tabular}{ccc}
   \includegraphics[width=0.45\linewidth]{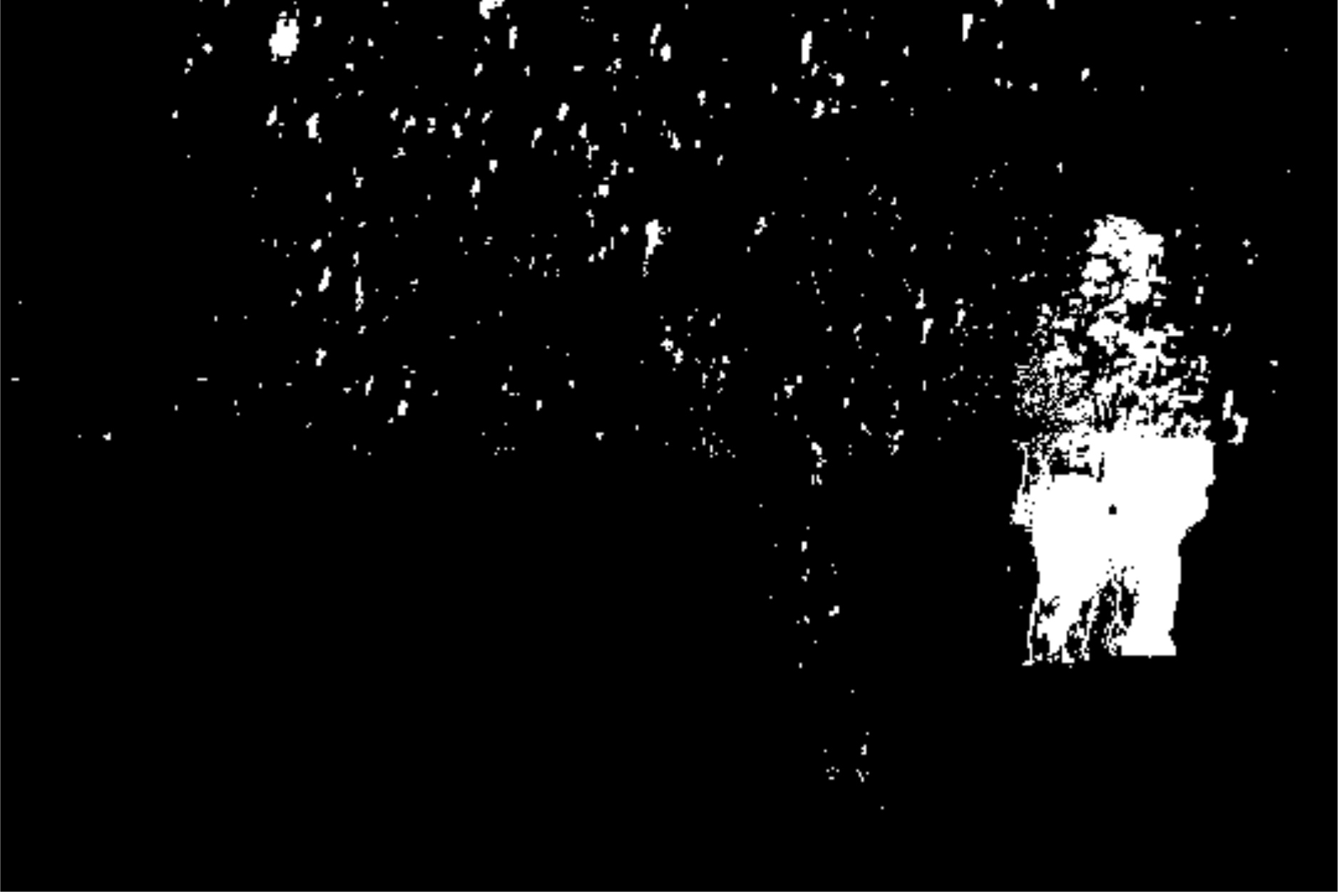}\label{fig:gmm_raw}&
	 \includegraphics[width=0.45\linewidth]{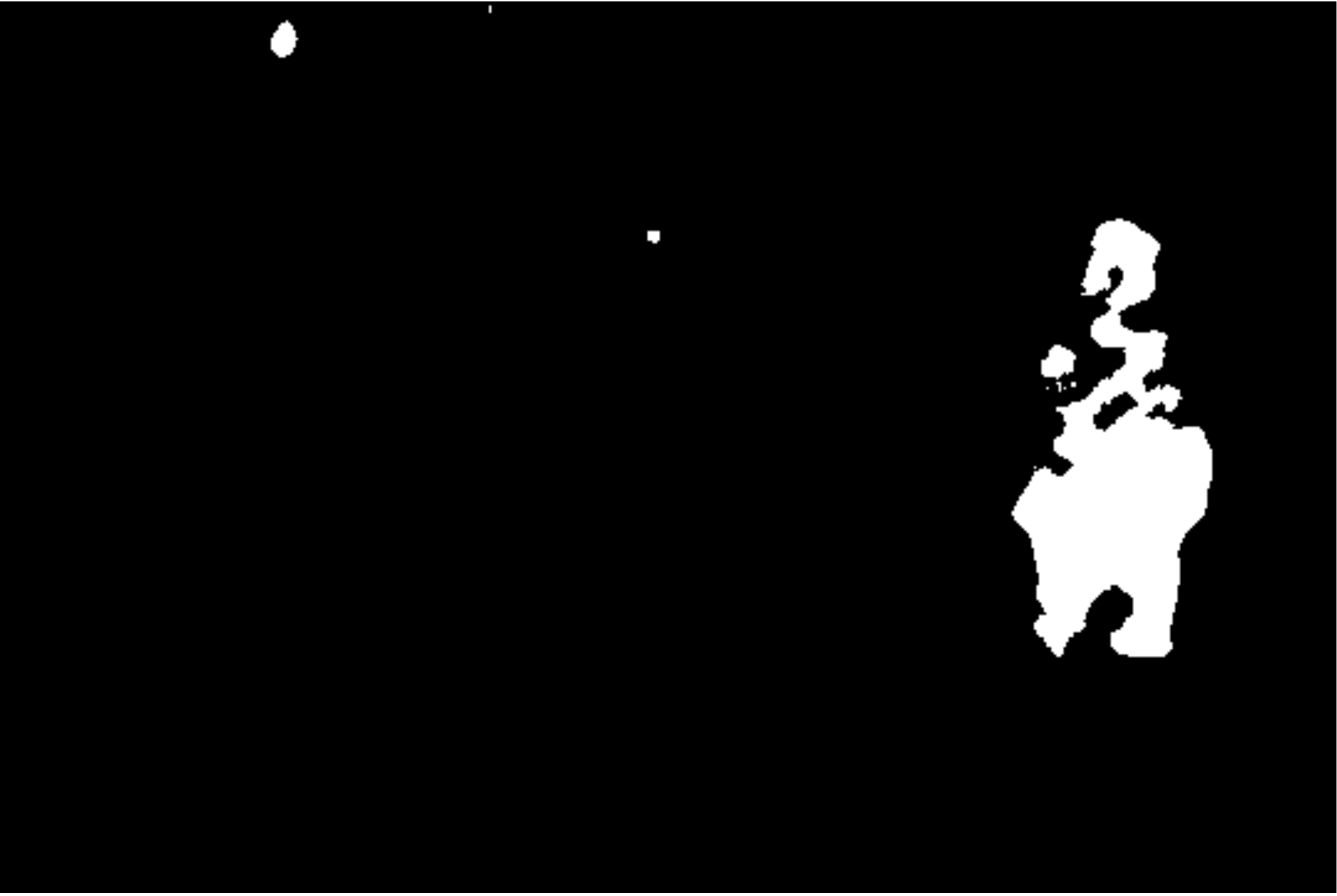}\label{fig:gmmmed}\\
	(a)&(b)\\
	\end{tabular}
   \caption[Effect of spatial-median filtering]{Effect of spatial-median filtering: (a) Raw segmentation result of the GMM algorithm \cite{stauffer1999mog}. (b) Spatial-median filtered output with a $9\times9$ kernel}
    \label{fig:pbas_raw_median}
\end{figure}

The spatial-median filtering, which is a commonly used post-processing method in background subtraction, returns the median over a neighborhood of given size (the kernel size) for each pixel in an image. As a consequence, the operation gets rid of outliers in the segmentation map and also performs blob smoothing (see Figure \ref{fig:pbas_raw_median}). After applying the spatial-median filter on the network output, we globally threshold the values for each pixel in order to map each pixel to $\{0,1\}$ . The threshold function is given by
\begin{equation}
g(z;R) = \begin{cases}
1 &\text{if $z > R$}\\
0 &\text{otherwise}
\end{cases},
\end{equation}
where \mbox{R} is the threshold level.

\section{Experiments}
\label{sec:experiments}

In order to evaluate our approach, we conducted several experiments on various data-sets. At first, we introduce the utilized data-sets for performance testing. Afterwards, the evaluation metrics are presented, which measure the quality of the segmentation outputs. The results on the evaluation data are subsequently reported. Furthermore, we analyze the network behavior during training. Additionally, we visualize the convolutional filters and generated feature maps at the end.

\subsection{Data-sets}
We employ multiple data-sets to perform our tests. The CDnet 2014 \cite{wang2014cdnet}, the Wallflower \cite{toyama1999wallflower} and the PETS 2009 data-set \cite{ferrymanpets2009}. The CDnet 2014 and also the Wallflower data-set were specifically designed for the background subtraction task. These data-sets contain video sequences from different categories, which correspond to the main challenges in background subtraction and also hand segmented ground truth images are also provided. The PETS 2009 \cite{ferrymanpets2009} data-set was designed for other purposes, such as the evaluation of tracking algorithms, and therefore no ground truth segmentations are available for its video sequences. The employed data-sets will be described in the following.

\subsubsection{CDnet 2014}
The CDnet 2014 \cite{wang2014cdnet} that was used for training of our CNNs will also be used for performance evaluation. With additional video sequences under new challenge categories, it is an extension of the CDnet 2012 \cite{goyette2012changedetection} data-set, which is the predecessor of the CDnet 2014 \cite{wang2014cdnet}.\\
For each video sequence in the data-set, corresponding ground truth segmentation images are available. For the newly added categories in CDnet 2014 \cite{wang2014cdnet} (see Table \ref{tab:cdnet2012_vs_cdnet2014}), only half of the ground truth segmentations were provided to avoid parameter tuning of background subtraction algorithms on the benchmark data-set. One has to refer to the online evaluation\footnote{\url{http://changedetection.net}} in order to get the results over all ground truth segmentations.

\begin{table}[]
\centering
\begin{tabular}{|c|c|c|c|}
\hline
\textbf{Categories}                                                  & \textbf{Video Sequences} & \textbf{CDnet 2012 \cite{goyette2012changedetection}} & \textbf{CDnet 2014 \cite{wang2014cdnet}} \\ \hline
badWeather                                                           & 4                        &                     & X                   \\ \hline
baseline                                                             & 4                        & X                   & X                   \\ \hline
cameraJitter                                                         & 4                        & X                   & X                   \\ \hline
dynamicBackground                                                    & 6                        & X                   & X                   \\ \hline
\begin{tabular}[c]{@{}c@{}}intermittent-\\ ObjectMotion\end{tabular} & 6                        & X                   & X                   \\ \hline
lowFramerate                                                         & 4                        &                     & X                   \\ \hline
nightVideos                                                          & 6                        &                     & X                   \\ \hline
PTZ                                                                  & 4                        &                     & X                   \\ \hline
shadow                                                               & 6                        & X                   & X                   \\ \hline
thermal                                                              & 5                        & X                   & X                   \\ \hline
turbulence                                                           & 4                        &                     & X                   \\ \hline
\end{tabular}
\caption[Comparison of CDnet 2012 \cite{goyette2012changedetection} and CDnet 2014 \cite{wang2014cdnet}]{Comparison of CDnet 2012 \cite{goyette2012changedetection} and CDnet 2014 \cite{wang2014cdnet}: Categories that are marked with a cross are contained in the respective data-set.}
\label{tab:cdnet2012_vs_cdnet2014}
\end{table}

\subsubsection{Wallflower}
Another data-set that we employ for evaluation purpose is the Wallflower data-set \cite{toyama1999wallflower}. For each category in the data-set, there exists a single video sequence (see Table \ref{wallflower_overview}). Also, for each video sequence, a hand segmented ground truth segmentation image is provided. Hence, when evaluating background subtraction algorithms on the Wallflower data-set \cite{toyama1999wallflower}, only a single ground truth segmentation is considered.

\begin{table}[t]
\centering
\begin{tabular}{|c|c|c|}
\hline
\textbf{Video sequences/categories} & \textbf{Number of frames} & \textbf{\begin{tabular}[c]{@{}c@{}}Groundtruth \\ frame number\end{tabular}} \\ \hline
Bootstrap                           & 3055                      & 353                                                                          \\ \hline
Camouflage                          & 294                       & 252                                                                          \\ \hline
ForegroundAperture                  & 2113                      & 490                                                                          \\ \hline
LightSwitch                         & 2715                      & 1866                                                                         \\ \hline
MovedObject                         & 1745                      & 986                                                                          \\ \hline
TimeOfDay                           & 5890                      & 1850                                                                         \\ \hline
WavingTrees                         & 287                       & 248                                                                          \\ \hline
\end{tabular}
\caption{Overview of the Wallflower data-set \cite{toyama1999wallflower}}
\label{wallflower_overview}
\end{table}

\subsubsection{PETS 2009}
\begin{figure}%
\centering
\begin{tabular}{ccccc}
\includegraphics[height=0.75in,width=1in]{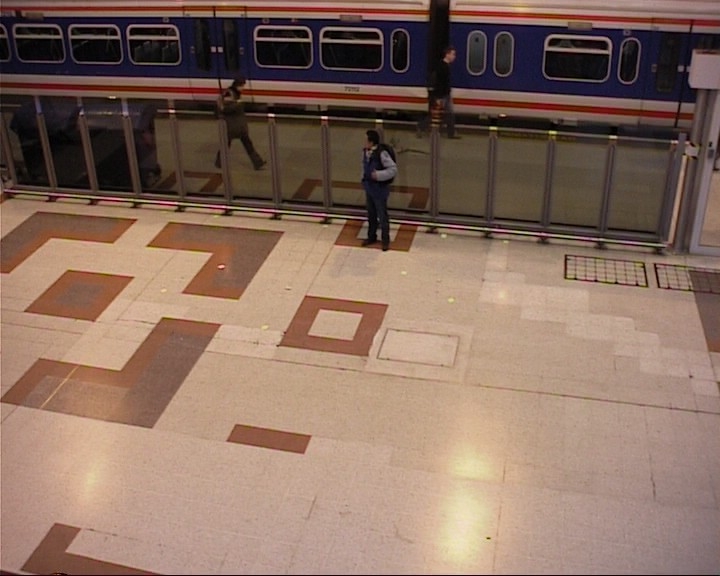}&\hspace{-0.1in}
\includegraphics[height=0.75in,width=1in]{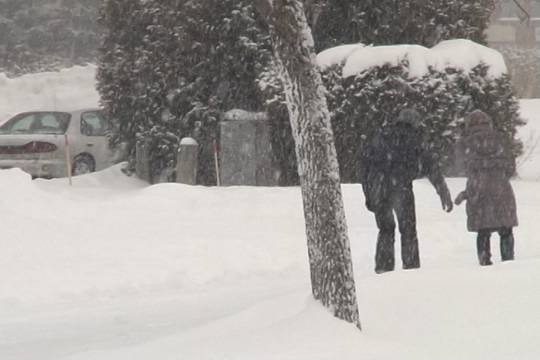}&\hspace{-0.1in}
\includegraphics[height=0.75in,width=1in]{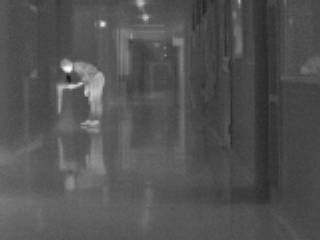}&\hspace{-0.1in}
\includegraphics[height=0.75in,width=1in]{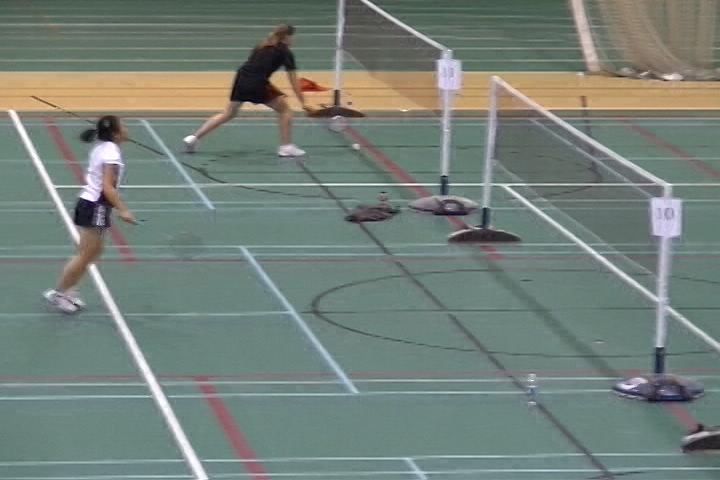}&\hspace{-0.1in}
\includegraphics[height=0.75in,width=1in]{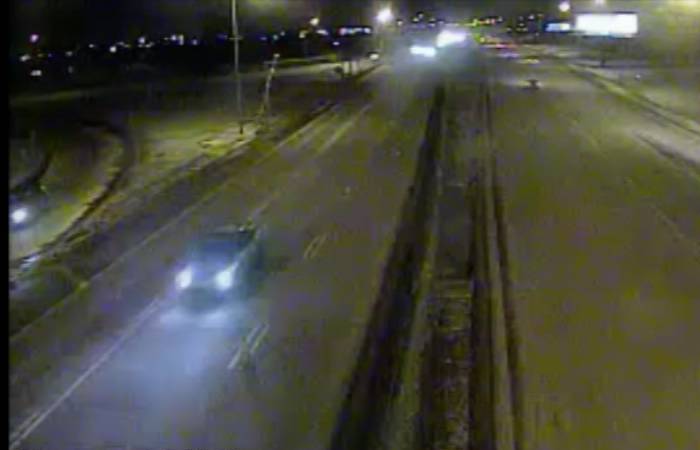}\\
(a)&(b)&(c)&(d)&(e)\\
\\
\includegraphics[height=0.75in,width=1in]{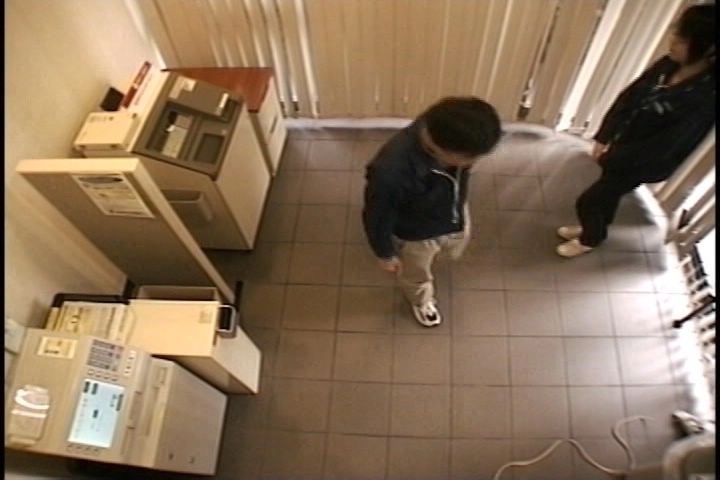} &\hspace{-0.1in}
\includegraphics[height=0.75in,width=1in]{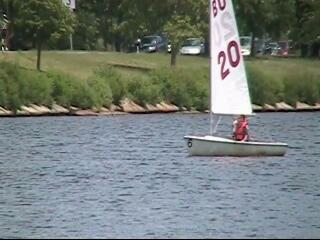} &\hspace{-0.1in}
\includegraphics[height=0.75in,width=1in]{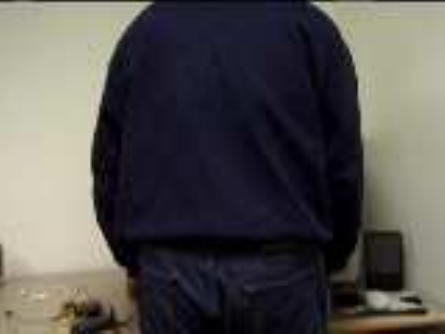} &\hspace{-0.1in}
\includegraphics[height=0.75in,width=1in]{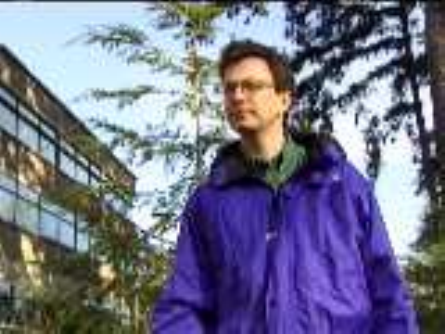}& \hspace{-0.1in}
\includegraphics[height=0.75in,width=1in]{figures/image_examples/view001_row/view001_frame_0146} \\ \hspace{-0.1in}
(f)&(g)&(h)&(i)&(j)

\end{tabular}
\caption{Sample images of each video sequence. (a) baseline; (b) bad weather; (c) thermal; (d) camera jitter; (e) night videos; (f) shadow; (g) dynamic background; (h) Camouflage \cite{toyama1999wallflower}; (i) WavingTrees \cite{toyama1999wallflower}; (j) PETS2009 \cite{ferrymanpets2009}      }%
\label{fig:selected_video_sequences}%
\end{figure}

The PETS 2009 data-set \cite{ferrymanpets2009} is a benchmark for tracking of individual people within a crowd. It consists of multiple video sequences, recorded from static cameras, and different crowd activities. Since the data-set is not designed for background subtraction evaluation, there are no ground truth segmentation images for this purpose. Thus, only the qualitative segmentation results will be evaluated without calculating any evaluation metric.
A sample image from each category is represented in Fig. \ref{fig:selected_video_sequences}. 

\subsection{Evaluation Metrics}
\label{sec:eval_metrics}
In order to measure the quality of a background subtraction algorithm, we evaluate the performance by comparing the output segmentations with the groundtruth segmentations to get the following statistics:\\
\\
\textbf{True Positive (TP)}: Foreground pixels in the output segmentation that are also foreground pixels in the ground truth segmentation.\\
\\
\textbf{False Positive (FP)}: Foreground pixels in the output segmentation that are not foreground pixels in the ground truth segmentation.\\
\\
\textbf{True Negative (TN)}: Background pixels in the output segmentation that are also background pixels in the ground truth segmentation.\\
\\
\textbf{False Negative (FN)}: Background pixels in the output segmentation that are not background pixels in the ground truth segmentation.\\
\\
Using these statistics, different evaluation metrics are calculated that are outlined in the following:

\begin{itemize}
	\item Recall (Re):
  \begin{equation}
  Re = \frac{TP}{TP + FN}
  \end{equation}
	\item Specificity (Sp):
  \begin{equation}
  Sp = \frac{TN}{TN + FP}
  \end{equation}
  
  \item Precision (Pr):
  \begin{equation}
  Pr = \frac{TP}{TP + FP}
  \end{equation}
	\item F Measure (FM):
  \begin{equation}
   FM= \frac{2 * Pr * Re}{Pr + Re}
  \end{equation}

	\item False Positive Rate (FPR):
  \begin{equation}
  FPR = \frac{FP}{FP + TN}
  \end{equation}
	\item False Negative Rate (FNR):
  \begin{equation}
    FNR = \frac{FN}{TP + FN}
  \end{equation}
  
  \item Percentage of Wrong Classifications (PWC):
  \begin{equation}
    PWC = 100 * \frac{FN + FP}{TP + FN + FP + TN}
   \end{equation}
\end{itemize}
We are especially interested in the FM metric since most state-of-the-art algorithms in background subtraction typically exhibit higher FM values than worse performing background subtraction algorithms. This is due to the combination of multiple evaluation metrics for the calculation of the FM. Thus, the overall performance of a background subtraction algorithm is highly coupled with its FM performance. 
\subsection{Evaluation Process}
In Section \ref{sec:approach}, we have introduced our background subtraction system, consisting of a CNN, a background image retrieval and a post-processing module. For the background image retrieval and the post-processing technique, respectively two methods were proposed. The novel background image generation based on the SuBSENSE \cite{st2015subsense} and Flux tensor was proposed and for the post-processing, spatial-median filtering is considered. 
In order to get the best performing setup, we calculate the evaluation metrics, presented in Section \ref{sec:eval_metrics}, over the CDnet 2014 \cite{wang2014cdnet}, for each setup of our background subtraction system. Also, we compare the category-wise FM and the overall FM for those data-sets with the FMs from other background subtraction algorithms. Afterward, we will select the best performing setup for further comparison. For the Wallflower data-set \cite{toyama1999wallflower}, we calculate the FM for the best setup and again, compare the values with those from other algorithms. Due to the missing ground truth images for the videos in the PETS 2009 data-set \cite{ferrymanpets2009}, we can not derive the numeric values for the FM and therefore only evaluate the segmentation outputs.
In order to compare our approach with other algorithms on multiple data-sets, we need to be able to generate the corresponding segmentation, using different algorithms. For the  CDnet 2014 \cite{wang2014cdnet}, all algorithms that are listed in the online ranking\footnote{\url{http://changedetection.net}}, their evaluation metrics and segmentation results are available. For the other data-sets, namely the PETS 2009 \cite{ferrymanpets2009} and the Wallflower data-set \cite{toyama1999wallflower}, we need to explicitly generate the segmentation images for those video sequences. For this purpose, we employed the BGSLibrary \cite{bgslibrary} as for the background image generation. As a consequence, we compare our method with certain algorithms that were online evaluated for the CDnet 2014 \cite{wang2014cdnet}, as well as implemented in the BGSLibary \cite{bgslibrary}.\\
\\
\subsubsection{Network Training}
\begin{figure}[t]
\centering
\begin{tabular}{c}
\includegraphics[width=0.47\textwidth]{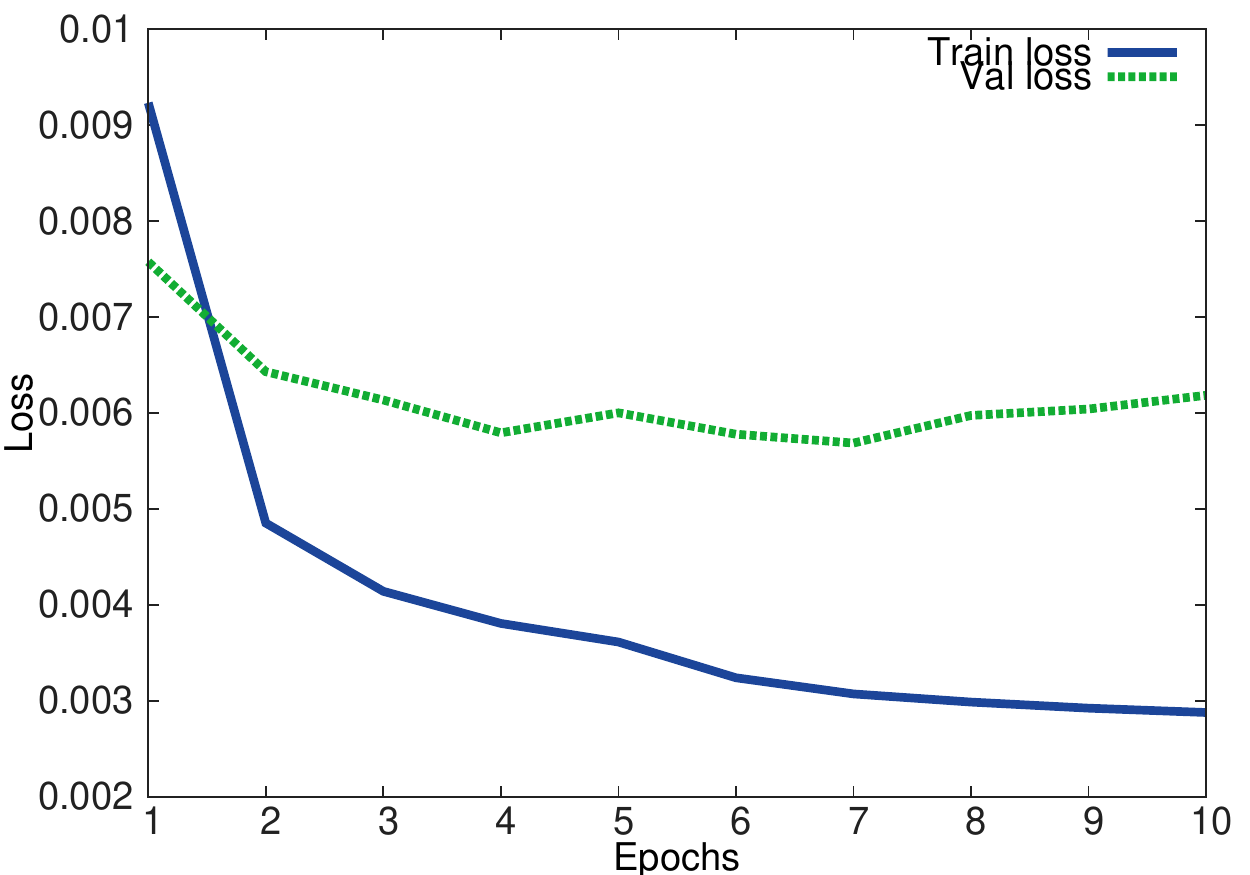}\\
(a)\\
\end{tabular}
\caption[Training and validation loss plot]{Training and validation loss of the network using background images generated using the proposed background model.}
\label{fig:network_training}
\end{figure}

As already mentioned in Section \ref{sec:approach}, we train the network with mini-batches of size 150, a learning rate $\alpha=2.5*10^{-3}$ over 10 epochs. The training data comprises of 150 images per video sequence in the categories listed in Table \ref{tab:cdnet2014_training_categories}. The validation data contains 20 frames per video sequence. We train the network with RMSprop \cite{hinton2012lecturermsprop} using the BCE for the loss function. In Figure \ref{fig:network_training}, the training plot is illustrated. Both networks yield a similar behavior and performance during training, mainly due to the identical network architecture and training setup.

\subsection{Results}
\label{sec:network_performance}
In order to measure the quality of our background subtraction algorithm, we compute the different evaluation metrics for our background subtraction system. The metrics are reported for the CDnet 2014 \cite{wang2014cdnet}. Also, we compare our FMs with the FMs from other background subtraction algorithms over the complete evaluation data. For this purpose, we compare our algorithm with the GMM \cite{stauffer1999mog}, PBAS \cite{hofmann2012pbas} and the SuBSENSE \cite{st2015subsense} algorithm. In addition, for both CDnet data-sets, we employ further background subtraction algorithms for the FM comparison. Also, we compare the segmenation outputs among the algorithms. For the comparison, we select further video sequences from the PETS 2009 data-set \cite{ferrymanpets2009}.





\subsubsection{CDnet 2014}
\label{sec:cdnet2014_evaL}
In the following, for each setup of our background subtraction system, the evaluation metrics for the CDnet 2014 are listed in Tables \ref{tab:subsense_bg_median_pp_14}. The outputs are post-processed with a median filter with a size of $9\times 9$. The outputs of the CNN are compared with the threshold at $R=0.3$. As we can see in Table \ref{tab:subsense_bg_median_pp_14}, the CNN yields very good results. In addition, the CNN yields the highest FM among the other algorithms in 6 out of 11 categories (see Table \ref{tab:fmeasure_cdnet2012} and Table \ref{tab:fmeasure_cdnet2014} ). 
\begin{table}[]
\centering
\resizebox{\textwidth}{70pt}{
\begin{tabular}{|l|c|c|c|c|c|c|c|c|}
\hline
Category                    & Re     & Sp     & FPR    & FNR    & PWC     & FM      & Pr  \\ \hline
Baseline                    & 0.9517 & 0.9991 & 0.0013 & 0.0483 & 0.2424  & 0.9580  & 0.9660 \\ 
Camera jitter               & 0.8788 & 0.9957 & 0.0043 & 0.1212 & 0.8994  & 0.8990  & 0.9313  \\ 
Dynamic Background   	      & 0.8543 & 0.9988 & 0.0012 & 0.1457 & 0.2067  & 0.8761  & 0.9083  \\
Intermittent Object Motion  & 0.5735 & 0.9949 & 0.0051 & 0.4265 & 4.1292  & 0.6098  & 0.8251 \\ 
Shadow                      & 0.9584 & 0.9942 & 0.0058 & 0.0416 & 0.7403  & 0.9304  & 0.4844 \\ 
Thermal                     & 0.6637 & 0.9956 & 0.0044 & 0.3363 & 3.5773  & 0.7583  & 0.9257  \\ 
Bad Weather         				& 0.7517 & 0.9996 & 0.0004 & 0.2483 & 0.3784  & 0.8301  & 0.9494 \\ 
Low Framerate       				& 0.5924 & 0.9975 & 0.0025 & 0.4076 & 1.3564  & 0.6002  & 0.9677 \\ 
Night Videos        				& 0.5315 & 0.9959 & 0.0041 & 0.4685 & 2.5754  & 0.5835  & 0.8366 \\ 
PTZ                 				& 0.7549 & 0.9248 & 0.0752 & 0.2541 & 7.7228  & 0.3133  & 0.2855 \\ 
Turbulence          				& 0.7979 & 0.9998 & 0.0002 & 0.2021 & 0.0838  & 0.8455  & 0.9082 \\ 
Overall                   	& 0.7545 & 0.9905 & 0.0095 & 0.2455 & 1.9920  & 0.7548  & 0.8332 \\ \hline
\end{tabular}}
\caption{Evaluation of CNN on the CDnet 2014}
\label{tab:subsense_bg_median_pp_14}
\end{table}

\begin{table}[!h]
\centering

\resizebox{1.0\textwidth}{70pt}{
\begin{tabular}{llllllll}
Method       & FM\textsubscript{BSL}     & FM\textsubscript{DBG}	& FM\textsubscript{CJT}              & FM\textsubscript{IOM}       & FM\textsubscript{SHD}       & FM\textsubscript{THM}             \\ \hline\\
CNN          & \textbf{0.9580} 		& \textbf{0.8761}  		& \textbf{0.8990} 		& 0.6098 		& \textbf{0.9304} 		& 0.7583		 	 \\

PBAS \cite{hofmann2012pbas}        	& 0.9242       	& 0.6829       	& 0.7220       	& 0.5745      	& 0.8143      	& 0.7556     	    \\

PAWCS \cite{st2015pawcs}       		& 0.9397     	& 0.8938      	& 0.8137       	& \textbf{0.7764}       	& 0.8913    	& \textbf{0.8324}      	    \\

SuBSENSE \cite{st2015subsense}     	& 0.9503       	& 0.8177       	& 0.8152      	& 0.6569      	& 0.8986       	& 0.8171       	     \\

MBS \cite{sajid2015colorspacessinglegaussian}       &  0.9287      	& 0.7915       	& 0.8367      	& 0.7568       	& 0.8262       	& 0.8194     		\\

GMM \cite{stauffer1999mog}        	& 0.8245       	& 0.6330      	& 0.5969      	& 0.5207       	& 0.7156      	& 0.6621      	    \\

RMoG \cite{varadarajan2015regionmog}        		& 0.7848       	& 0.7352     	& 0.7010      	& 0.5431     	& 0.7212      	& 0.4788            \\

Spectral-360 \cite{sedky2010image} 	& 0.9330      	& 0.7872      	& 0.7156      	& 0.5656     	& 0.8843     	& 0.7764     	   
\end{tabular}}
\caption{FM metric comparison of different background subtraction algorithms over the 6 categories of the CDnet2014, namely 1) Baseline (BSL), Dynamic background (DBG), Camera jitter (CJT), Intermittet objectmotion (IOM), Shadow (SHD),  Thermal (THM).}
\label{tab:fmeasure_cdnet2012}
\end{table}
\vspace{.2in}
\begin{table}[!t]
\centering
\resizebox{.9\textwidth}{70pt}{
\begin{tabular}{lllllll}
Method       & FM\textsubscript{BDW}     & FM\textsubscript{LFR}	& FM\textsubscript{NVD}              & FM\textsubscript{PTZ}       & FM\textsubscript{TBL}   \\ \hline\\
CNN          & 0.8301		& 0.6002		& \textbf{0.5835}  		& 0.3133 		& \textbf{0.8455} 		  	 \\

PBAS \cite{hofmann2012pbas}        	& 0.7673      	& 0.5914      	& 0.4387       	& 0.1063       	& 0.6349       	    \\

PAWCS \cite{st2015pawcs}       		& 0.8152    	& \textbf{0.6588}     	& 0.4152       	& 0.4615       	& 0.6450     	     \\

SuBSENSE \cite{st2015subsense}     	& \textbf{0.8619}       	& 0.6445       	& 0.5599       	& 0.3476      	& 0.7792       	     \\

MBS \cite{sajid2015colorspacessinglegaussian}       & 0.7730      	& 0.6279       	& 0.5158       	& \textbf{0.5118}       	& 0.5698     	\\

GMM \cite{stauffer1999mog}        	& 0.7380       	& 0.5373      	& 0.4097       	& 0.1522       	& 0.4663     	   \\

RMoG \cite{varadarajan2015regionmog}        		& 0.6826       	& 0.5312      	& 0.4265      	& 0.2400     	& 0.4578           \\

Spectral-360 \cite{sedky2010image} 	& 0.7569      	& 0.6437      	& 0.4832      	& 0.3653     	& 0.5429     	  
\end{tabular}}
\caption{FM metric comparison of different background subtraction algorithms over the 5 categories of the CDnet2014, namely Bad weather (BDW), Low framerate (LFM), Night videos (NVD), PTZ (PTZ), Turbulence (TBL) }
\label{tab:fmeasure_cdnet2014}
\end{table}

\subsubsection{Wallflower Evaluation}
For the the Wallflower data-set \cite{toyama1999wallflower}, we compare the FMs among the considered algorithms. For each video in the data-set, only a single groundtruth image is given, therefore, the FM is calculated only for this ground truth image.\\
The different FMs are reported in Table \ref{tab:fmeasure_wallflower}. Please note that we do not consider the ''MovedObject'' video for comparison since the corresponding ground truth image contains no foreground pixels and therefore, the FM can not be calculated.\\
From the results we can see that the CNN yields the best overall FM among the considered background subtraction algorithms.\\

\begin{table}[!t]
\centering
\resizebox{.95\textwidth}{60pt}{
\begin{tabular}{lcccc}

FM\textsubscript{Video}                  	&  CNN & SuBSENSE \cite{st2015subsense} & PBAS \cite{hofmann2012pbas} & GMM \cite{stauffer1999mog}   \\ \hline
FM\textsubscript{Bootstrap}            	& \textbf{0.7479} & 0.4192   & 0.2857    & 0.5306 \\
FM\textsubscript{Camouflage}          		& \textbf{0.9857} & 0.9535   & 0.8922    & 0.8307 \\
FM\textsubscript{ForegroundAperture}       & 0.6583 & \textbf{0.6635}   & 0.6459    & 0.5778 \\
FM\textsubscript{LightSwitch}             	& \textbf{0.6114} & 0.3201   & 0.2212    & 0.2296 \\
FM\textsubscript{TimeOfDay}             	& 0.5494 & 0.7107   & 0.4875    & \textbf{0.7203} \\
FM\textsubscript{WavingTrees}    			& 0.9546 & 0.9597   & 0.8421    & \textbf{0.9767} \\ \hline
Overall    										& \textbf{0.7512} & 0.6711   & 0.5624    & 0.6443 \\ \hline
\end{tabular}}
\caption{F-Measures for the Wallflower dataset \cite{toyama1999wallflower}}
\label{tab:fmeasure_wallflower}
\end{table}

\begin{figure}[t]
\centering
\begin{tabular}{cccccc}
\hspace{-0.2in} \includegraphics[height=45pt,width=65pt]{figures/image_examples/pets_row/pets2006_in000560.jpg} & \hspace{-.01in}
\hspace{-0.2in} \includegraphics[height=45pt,width=65pt]{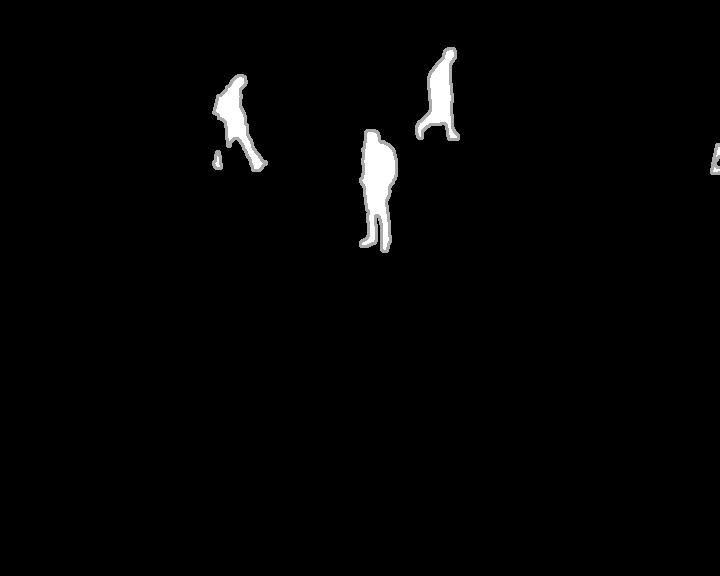} & \hspace{-.01in}
\hspace{-0.2in} \includegraphics[height=45pt,width=65pt]{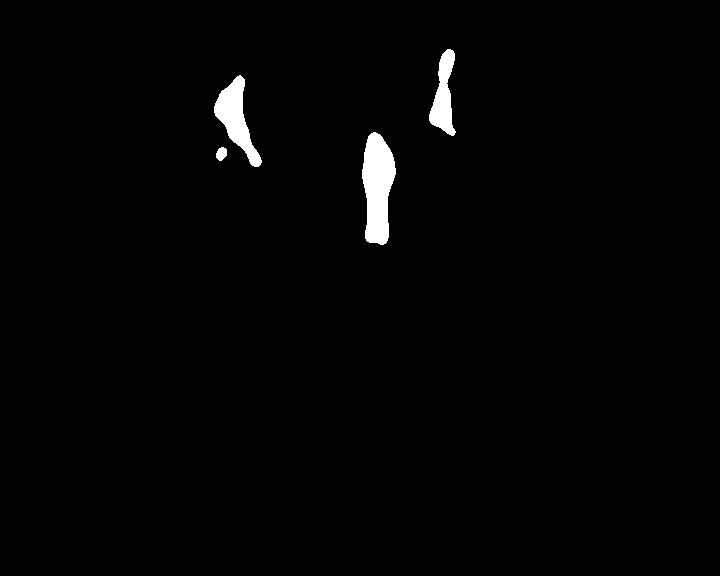} & \hspace{-.01in}
\hspace{-0.2in} \includegraphics[height=45pt,width=65pt]{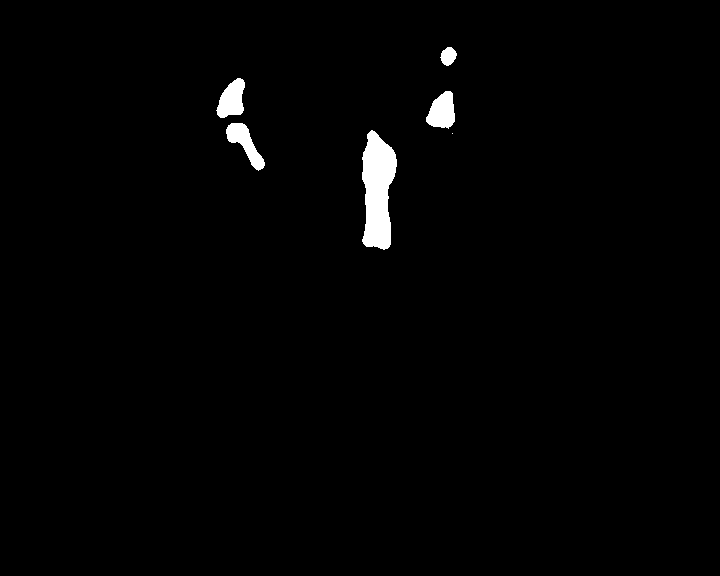}&  \hspace{-.01in}
\hspace{-0.2in} \includegraphics[height=45pt,width=65pt]{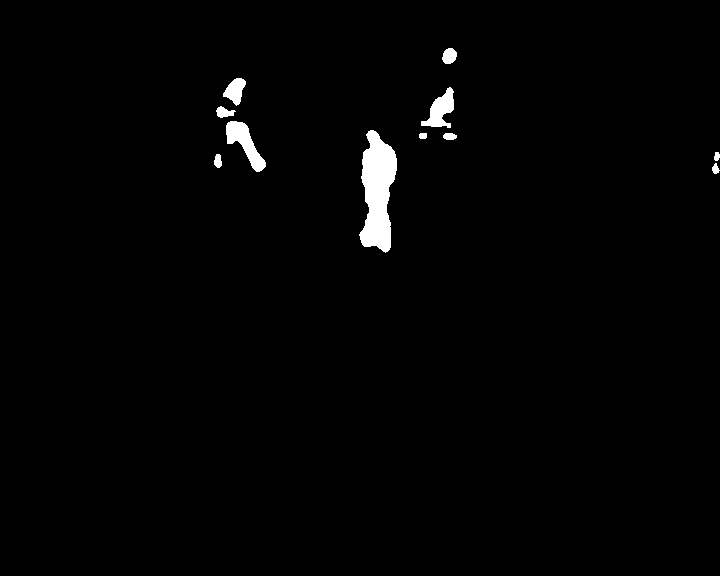} &\hspace{-.01in}
\hspace{-0.2in} \includegraphics[height=45pt,width=65pt]{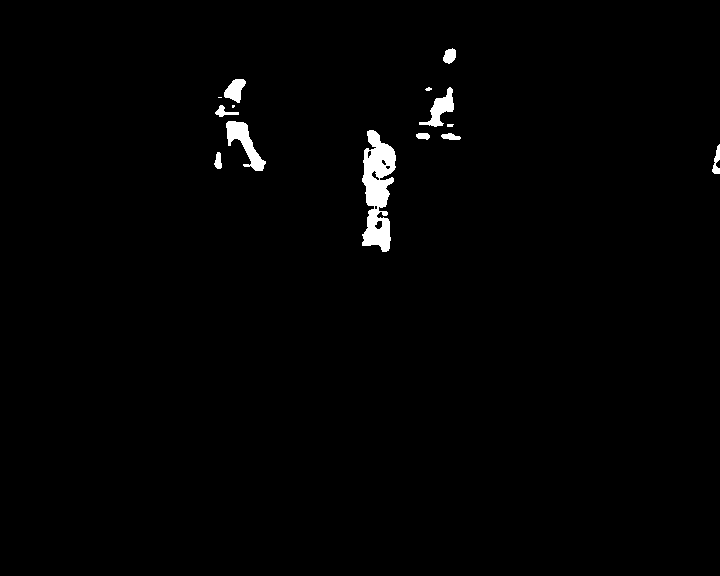} \\

\hspace{-0.2in} \includegraphics[height=45pt,width=65pt]{figures/image_examples/skating_row/skating_in000978.jpg} & \hspace{-.01in}
\hspace{-0.2in} \includegraphics[height=45pt,width=65pt]{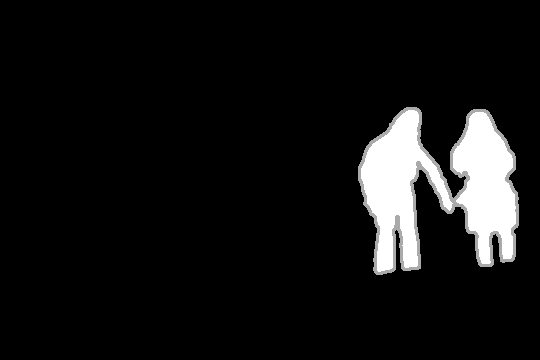} & \hspace{-.01in}
\hspace{-0.2in} \includegraphics[height=45pt,width=65pt]{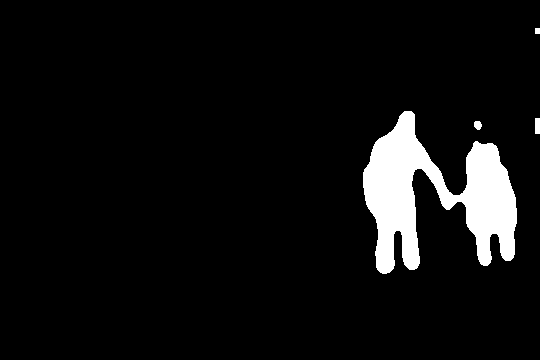} & \hspace{-.01in}
\hspace{-0.2in} \includegraphics[height=45pt,width=65pt]{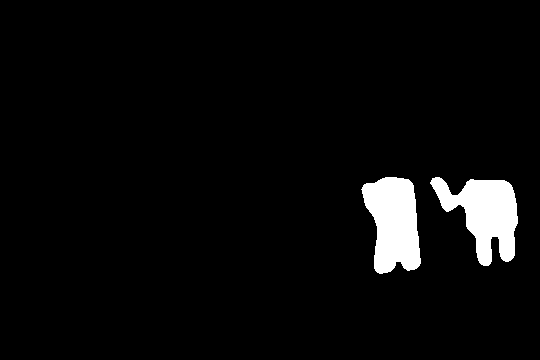}&  \hspace{-.01in}
\hspace{-0.2in} \includegraphics[height=45pt,width=65pt]{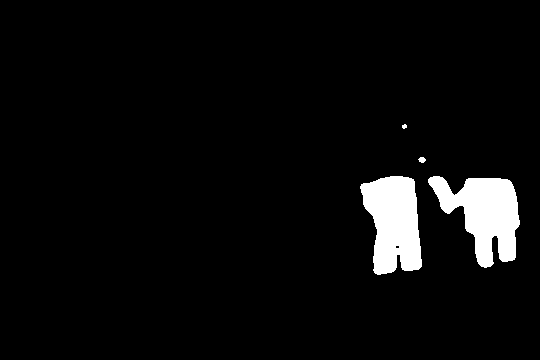} &\hspace{-.01in}
\hspace{-0.2in} \includegraphics[height=45pt,width=65pt]{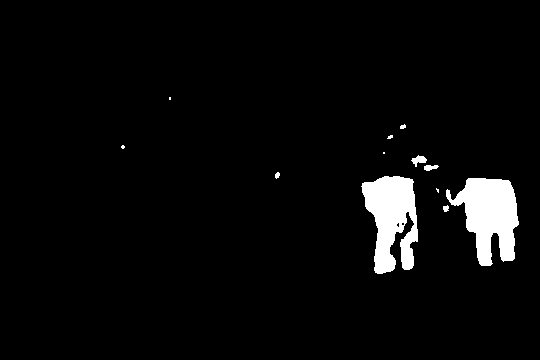} \\

\hspace{-0.2in} \includegraphics[height=45pt,width=65pt]{figures/image_examples/corridor_row/corridor_in001143} & \hspace{-.01in}
\hspace{-0.2in} \includegraphics[height=45pt,width=65pt]{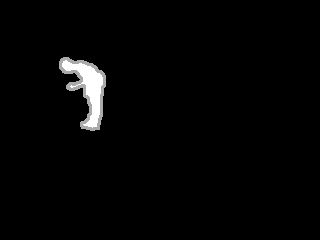} & \hspace{-.01in}
\hspace{-0.2in} \includegraphics[height=45pt,width=65pt]{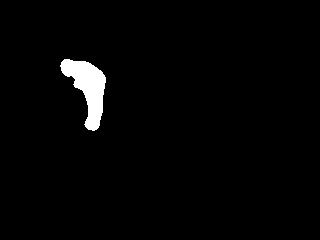} & \hspace{-.01in}
\hspace{-0.2in} \includegraphics[height=45pt,width=65pt]{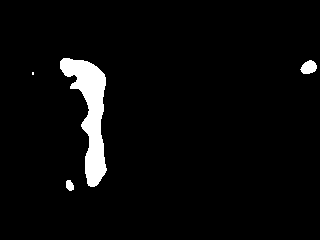}  & \hspace{-.01in}
\hspace{-0.2in} \includegraphics[height=45pt,width=65pt]{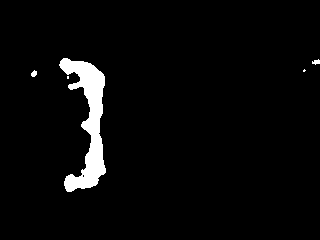}& \hspace{-.01in}
\hspace{-0.2in} \includegraphics[height=45pt,width=65pt]{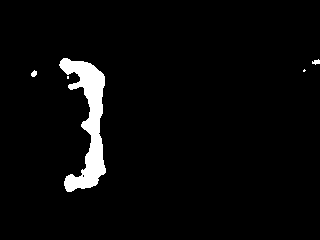}  \\

\hspace{-0.2in} \includegraphics[height=45pt,width=65pt]{figures/image_examples/badminton_row/badminton_in000888}  & \hspace{-.01in}
\hspace{-0.2in} \includegraphics[height=45pt,width=65pt]{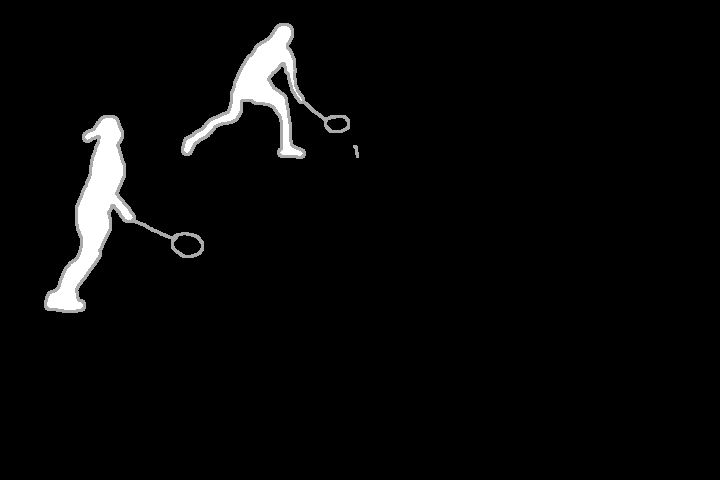}  & \hspace{-.01in}
\hspace{-0.2in} \includegraphics[height=45pt,width=65pt]{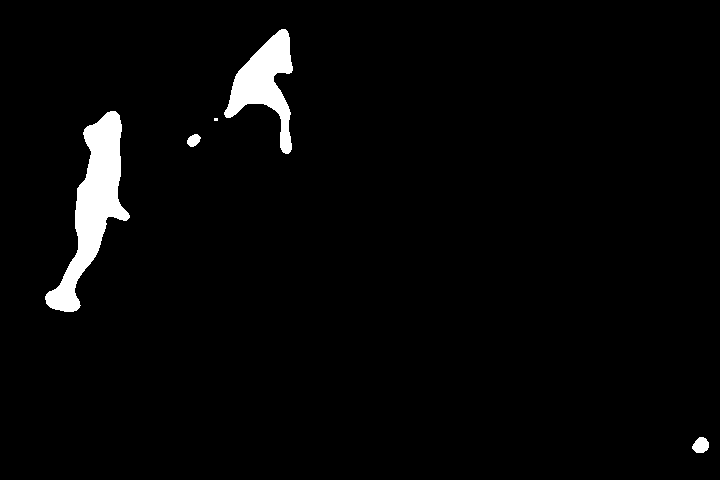}  & \hspace{-.01in}
\hspace{-0.2in} \includegraphics[height=45pt,width=65pt]{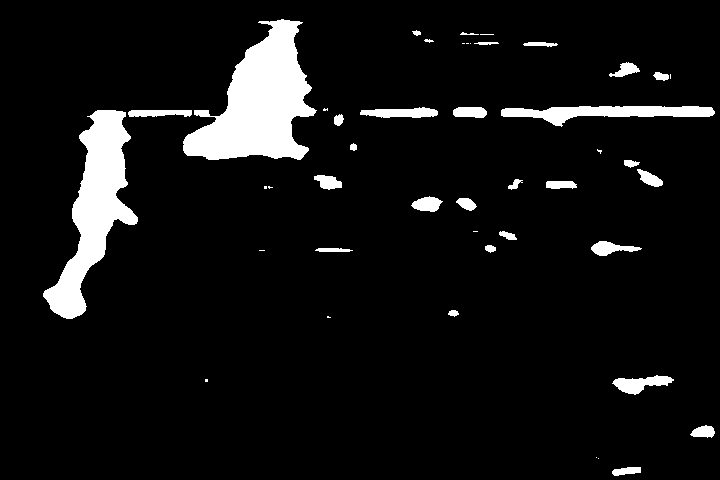}  & \hspace{-.01in} 
\hspace{-0.2in} \includegraphics[height=45pt,width=65pt]{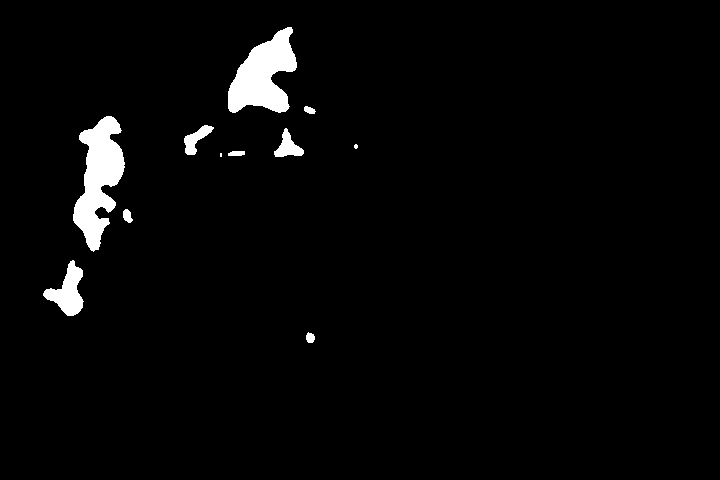}  & \hspace{-.01in}
\hspace{-0.2in} \includegraphics[height=45pt,width=65pt]{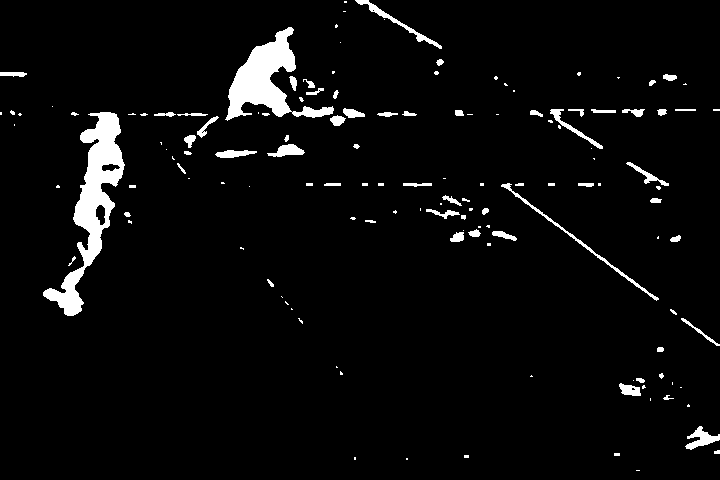} \\        

\hspace{-0.2in} \includegraphics[height=45pt,width=65pt]{figures/image_examples//fluidHighway_row/fluidHighway_in000629} & \hspace{-.01in}
\hspace{-0.2in} \includegraphics[height=45pt,width=65pt]{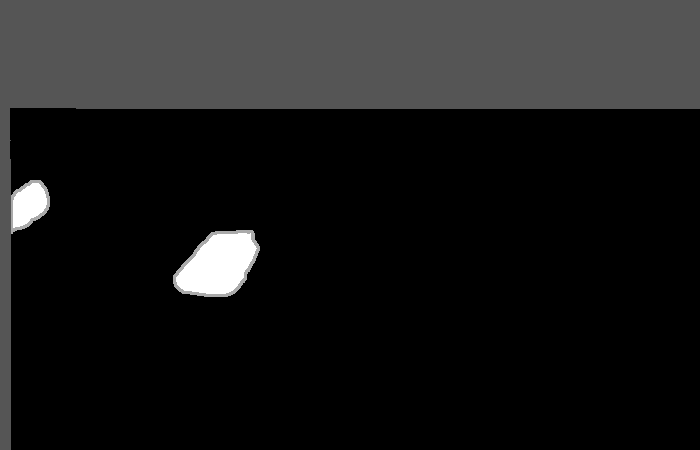}& \hspace{-.01in} 
\hspace{-0.2in} \includegraphics[height=45pt,width=65pt]{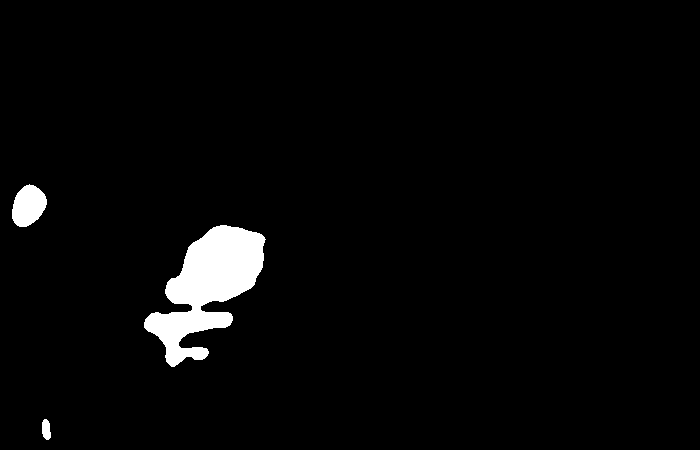}& \hspace{-.01in} 
\hspace{-0.2in} \includegraphics[height=45pt,width=65pt]{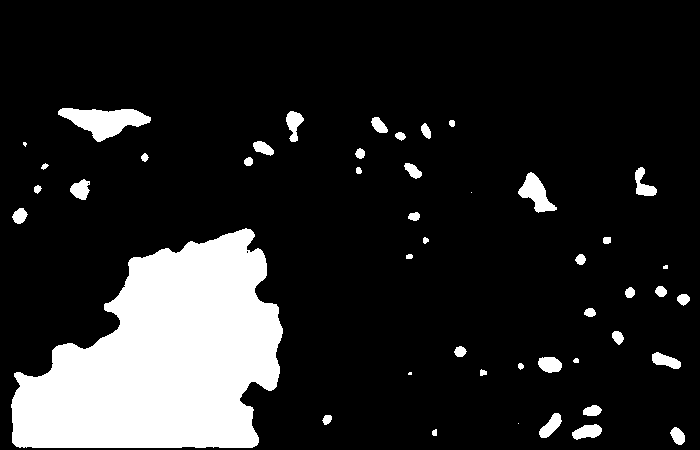}& \hspace{-.01in}  
\hspace{-0.2in} \includegraphics[height=45pt,width=65pt]{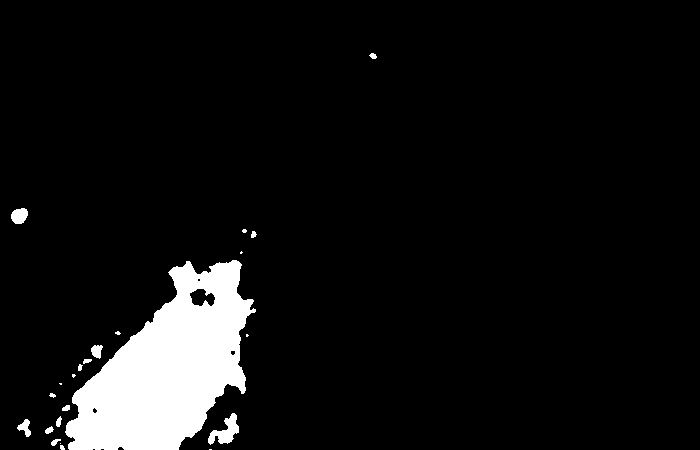}& \hspace{-.01in} 
\hspace{-0.2in} \includegraphics[height=45pt,width=65pt]{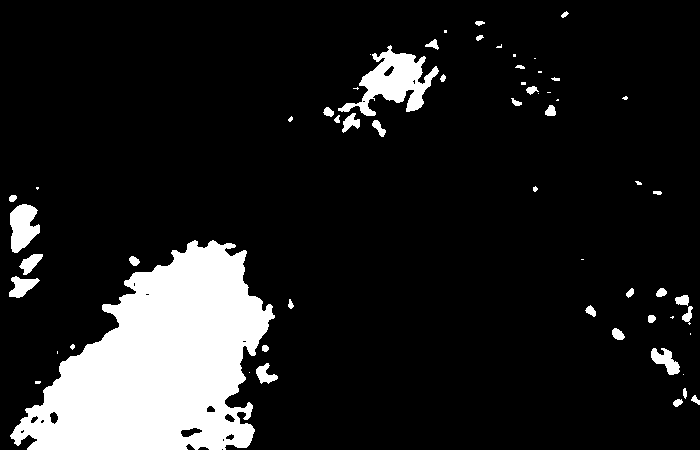}\\

\hspace{-0.2in} \includegraphics[height=45pt,width=65pt]{figures/image_examples/copyMachine_row/copyMachine_in001078}& \hspace{-.01in} 
\hspace{-0.2in} \includegraphics[height=45pt,width=65pt]{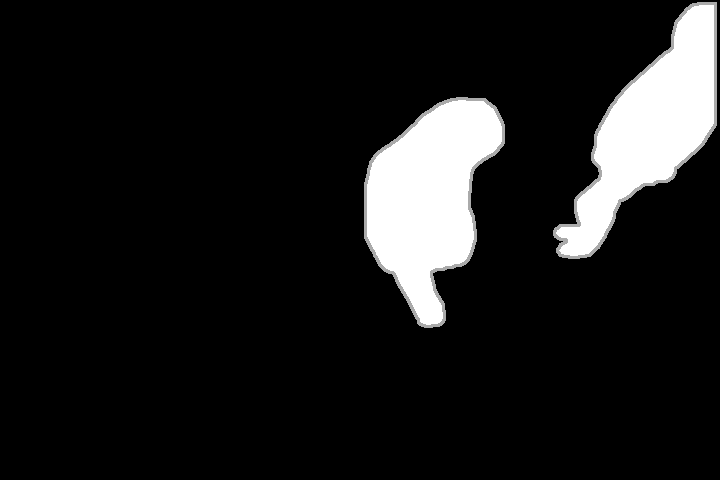}& \hspace{-.01in} 
\hspace{-0.2in} \includegraphics[height=45pt,width=65pt]{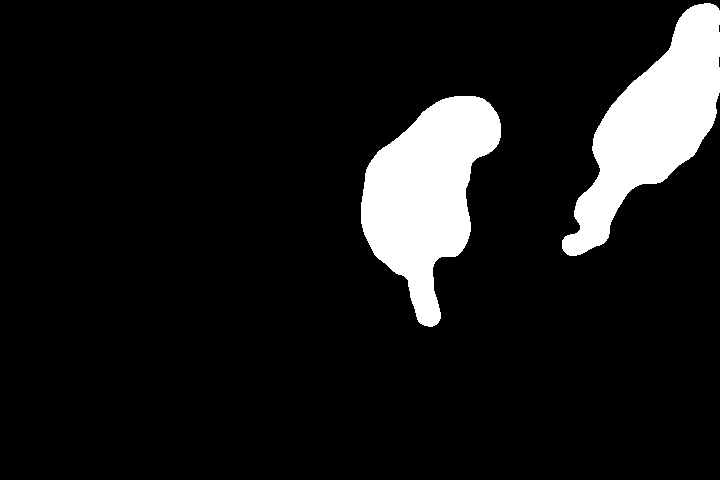}& \hspace{-.01in} 
\hspace{-0.2in} \includegraphics[height=45pt,width=65pt]{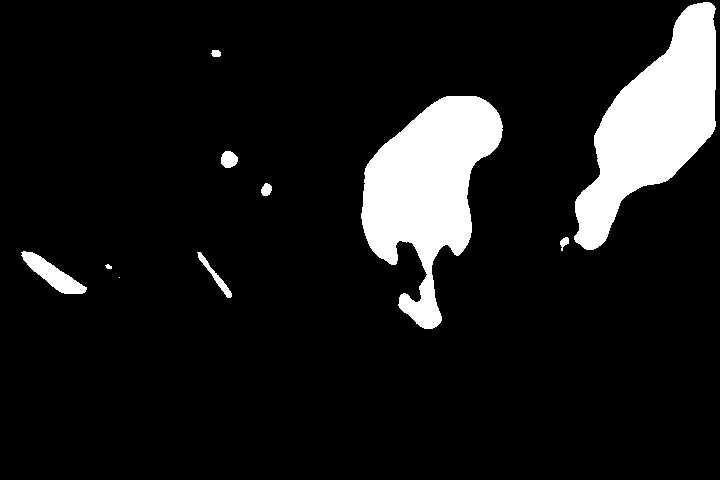}& \hspace{-.01in}  
\hspace{-0.2in} \includegraphics[height=45pt,width=65pt]{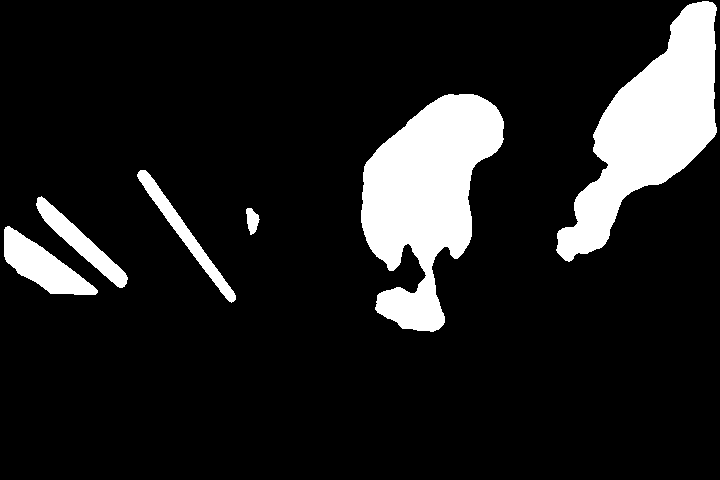}& \hspace{-.01in} 
\hspace{-0.2in} \includegraphics[height=45pt,width=65pt]{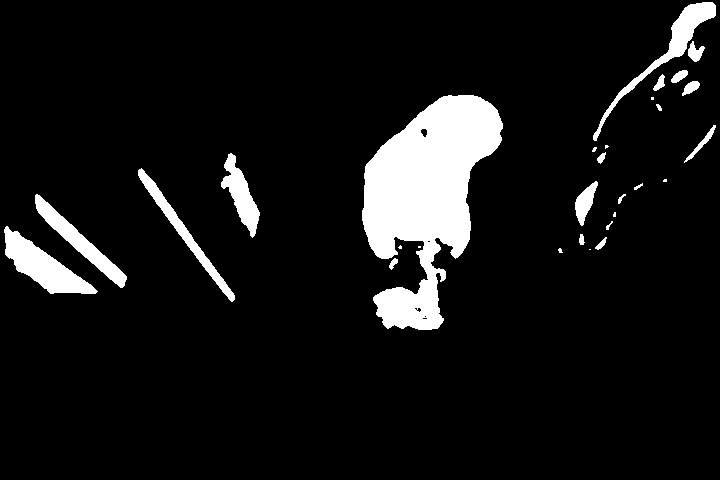}\\

\hspace{-0.2in} \includegraphics[height=45pt,width=65pt]{figures/image_examples/boats_row/boats_in007095}& \hspace{-.01in}  
\hspace{-0.2in} \includegraphics[height=45pt,width=65pt]{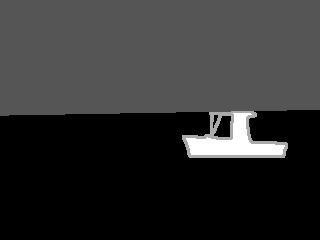}& \hspace{-.01in}  
\hspace{-0.2in} \includegraphics[height=45pt,width=65pt]{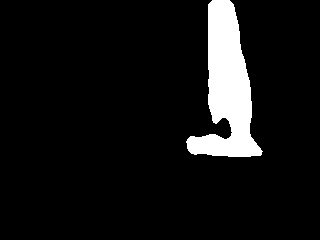}& \hspace{-.01in}  
\hspace{-0.2in} \includegraphics[height=45pt,width=65pt]{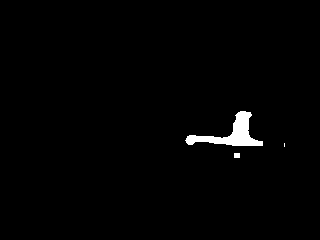}& \hspace{-.01in}   
\hspace{-0.2in} \includegraphics[height=45pt,width=65pt]{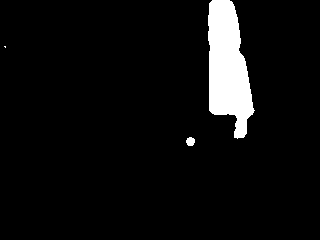}& \hspace{-.01in}  
\hspace{-0.2in} \includegraphics[height=45pt,width=65pt]{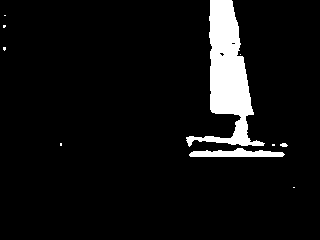}\\

\hspace{-0.2in} \includegraphics[height=45pt,width=65pt]{figures/image_examples/camouflage_row/camouflage_b00251}& \hspace{-.01in} 
\hspace{-0.2in} \includegraphics[height=45pt,width=65pt]{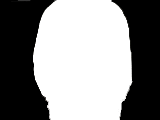}& \hspace{-.01in} 
\hspace{-0.2in} \includegraphics[height=45pt,width=65pt]{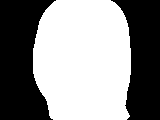}& \hspace{-.01in} 
\hspace{-0.2in} \includegraphics[height=45pt,width=65pt]{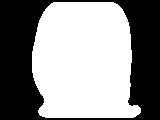}& \hspace{-.01in}  
\hspace{-0.2in} \includegraphics[height=45pt,width=65pt]{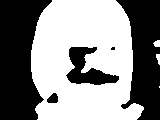}& \hspace{-.01in} 
\hspace{-0.2in} \includegraphics[height=45pt,width=65pt]{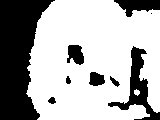}\\

\hspace{-0.2in} \includegraphics[height=45pt,width=65pt]{figures/image_examples/wavingtrees_row/waving_trees_b00247}& \hspace{-.01in} 
\hspace{-0.2in} \includegraphics[height=45pt,width=65pt]{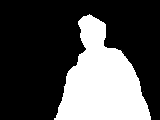}& \hspace{-.01in}
\hspace{-0.2in} \includegraphics[height=45pt,width=65pt]{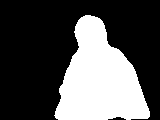}& \hspace{-.01in} 
\hspace{-0.2in} \includegraphics[height=45pt,width=65pt]{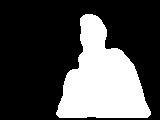}& \hspace{-.01in}  
\hspace{-0.2in} \includegraphics[height=45pt,width=65pt]{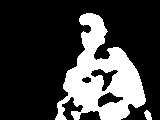}& \hspace{-.01in} 
\hspace{-0.2in} \includegraphics[height=45pt,width=65pt]{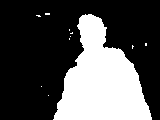}\\

\hspace{-0.2in} \includegraphics[height=45pt,width=65pt]{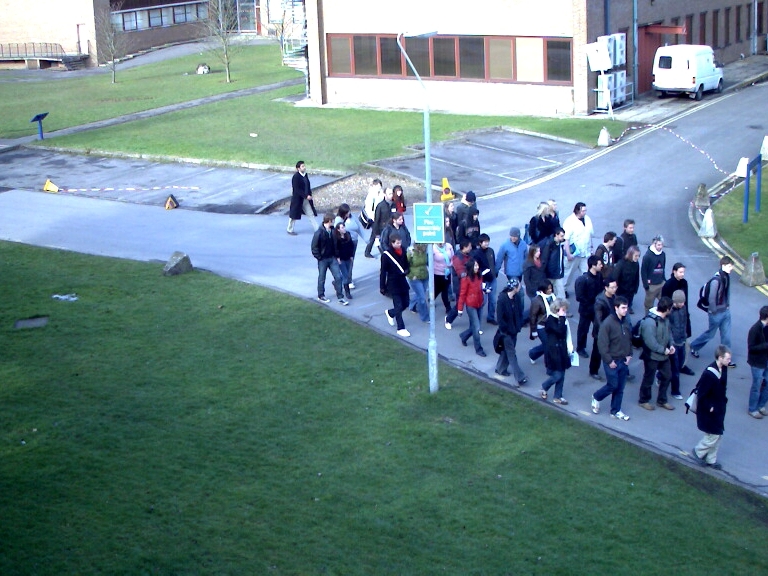} & \hspace{-.01in} 
\hspace{-0.2in} \includegraphics[height=45pt,width=65pt]{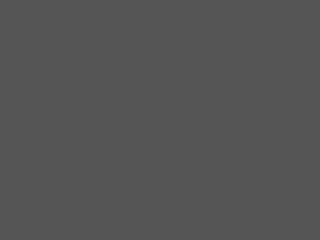} & \hspace{-.01in} 
\hspace{-0.2in} \includegraphics[height=45pt,width=65pt]{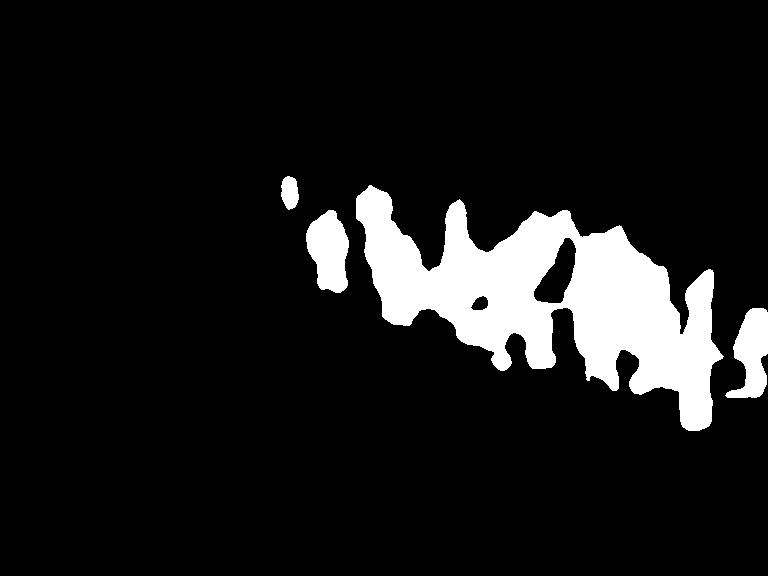} & \hspace{-.01in} 
\hspace{-0.2in} \includegraphics[height=45pt,width=65pt]{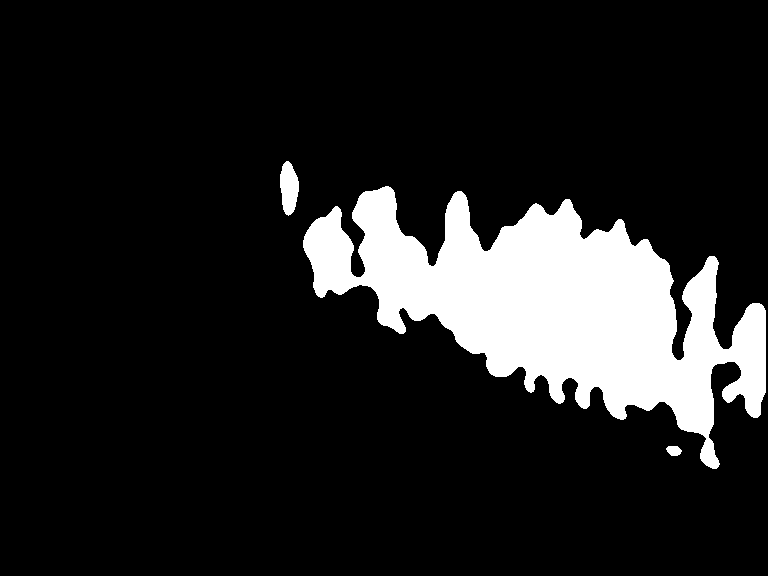}& \hspace{-.01in}   
\hspace{-0.2in} \includegraphics[height=45pt,width=65pt]{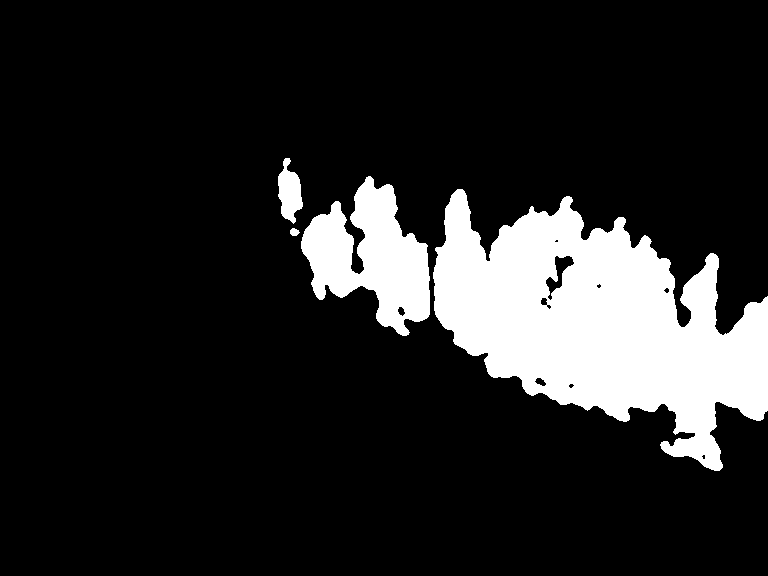} & \hspace{-.01in} 
\hspace{-0.2in} \includegraphics[height=45pt,width=65pt]{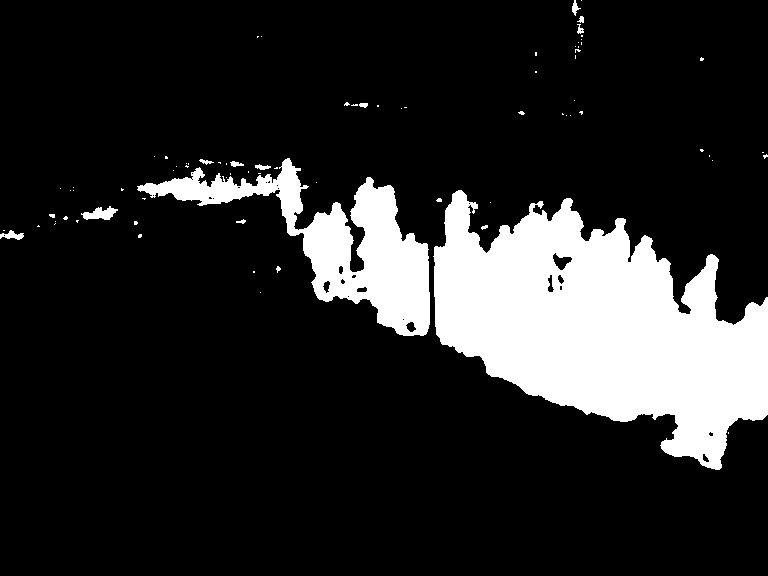}  \\

%

\end{tabular}
\caption{Comparison of the segmentation outputs: The first column is the input image, the second column is its ground truth image, the third column represents the output of the CNN. The fourth column is the output of the SuBSENSE \cite{st2015subsense}, the fifth column represents the output of the PBAS \cite{hofmann2012pbas} and the last column shows the output of the GMM \cite{stauffer1999mog} segmentation. Gray pixels in the ground truth segmentation indicate pixels which are not of interest}%
\label{tab:segmentation_compare}%
\end{figure}

\begin{figure}[th]
\centering
\begin{tabular}{ccc}
\includegraphics[height=70pt,width=0.3\textwidth]{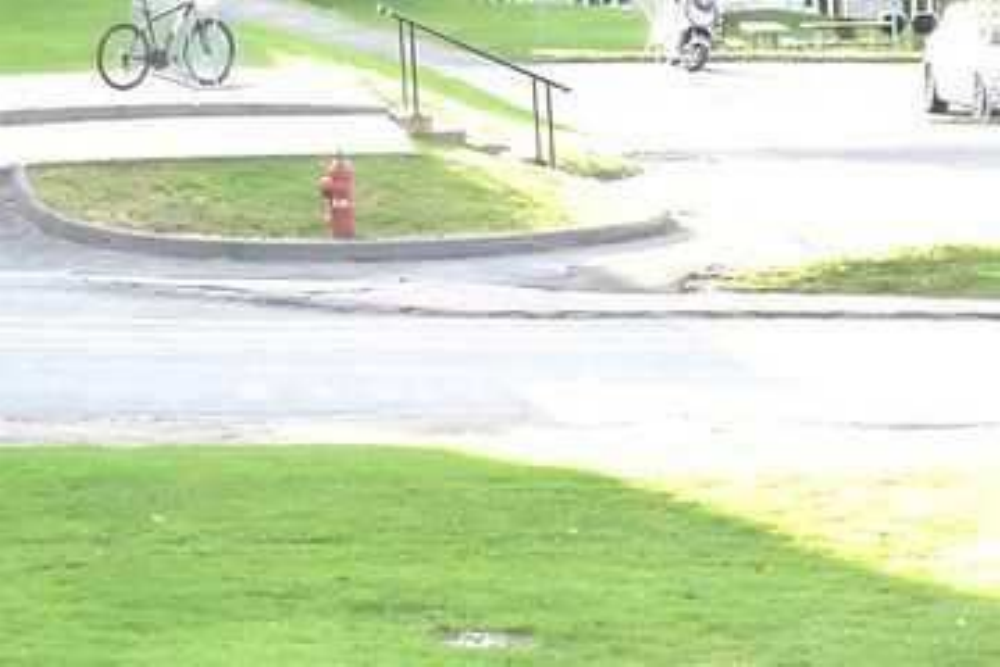}&
\includegraphics[height=70pt,width=0.3\textwidth]{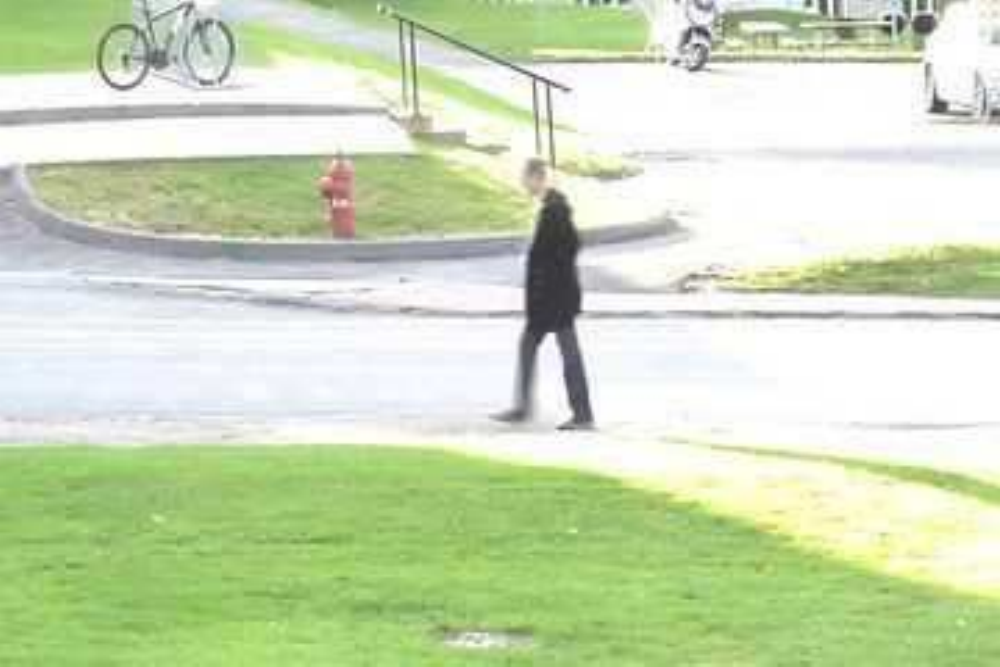}&
\includegraphics[height=70pt,width=0.3\textwidth]{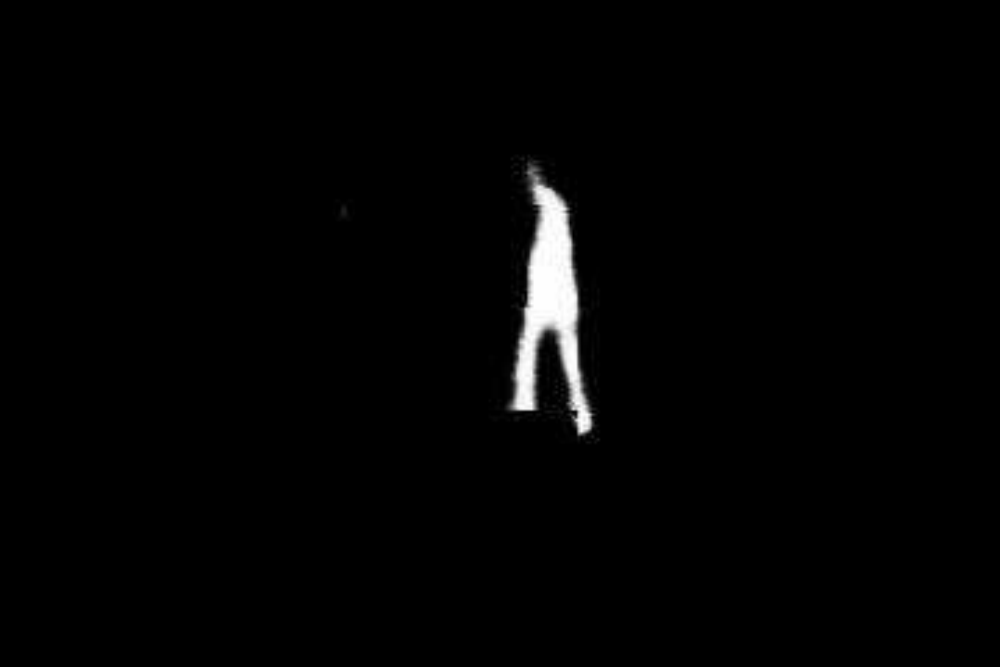}\\
(a)&(b)&(c)
\end{tabular}
\caption{Network input and output: (a) Background image of the input pair. (b) Input frame. (c) Network output.}
\label{fig:network_input_output}
\end{figure}

%

\subsection{Discussion}
The output segmentations are already precise and not very prone to outliers (e.g. through dynamic background regions), the main problems are the camouflage regions within foreground objects. 
In the first 6 categories of CDnet 2014, where mainly the feature extraction and segmentation are challenging, the CNN gives the best results, compared with other algorithms, in 6 out of 11 categories. For categories such as the ''PTZ (Pan Tilt Zoom)'' or ''low framerate'' category, the CNN yields poor results since the background images provided by the proposed algorithm are insufficient for performing good background subtraction. Therefore, our method gives poor segmentation for these categories and as a consequence, the average FM drops significantly.\\
However, for Wallflower data-set \cite{toyama1999wallflower}, where in most cases the background model is not challenged, our algorithm outperforms all other algorithms in terms of segmentation quality and overall FM.\\
To sum up, our system outperforms the existing algorithms when the challenge does not lie in the background modeling/maintenance, but on the other hand, due to the corruption of the background images, our method performs poorly once there are large changes in the background scenery.\\\\


\subsubsection{Network Analysis}
\begin{figure}[th]
\centering
\begin{tabular}{cc}
\includegraphics[width=0.4\textwidth]{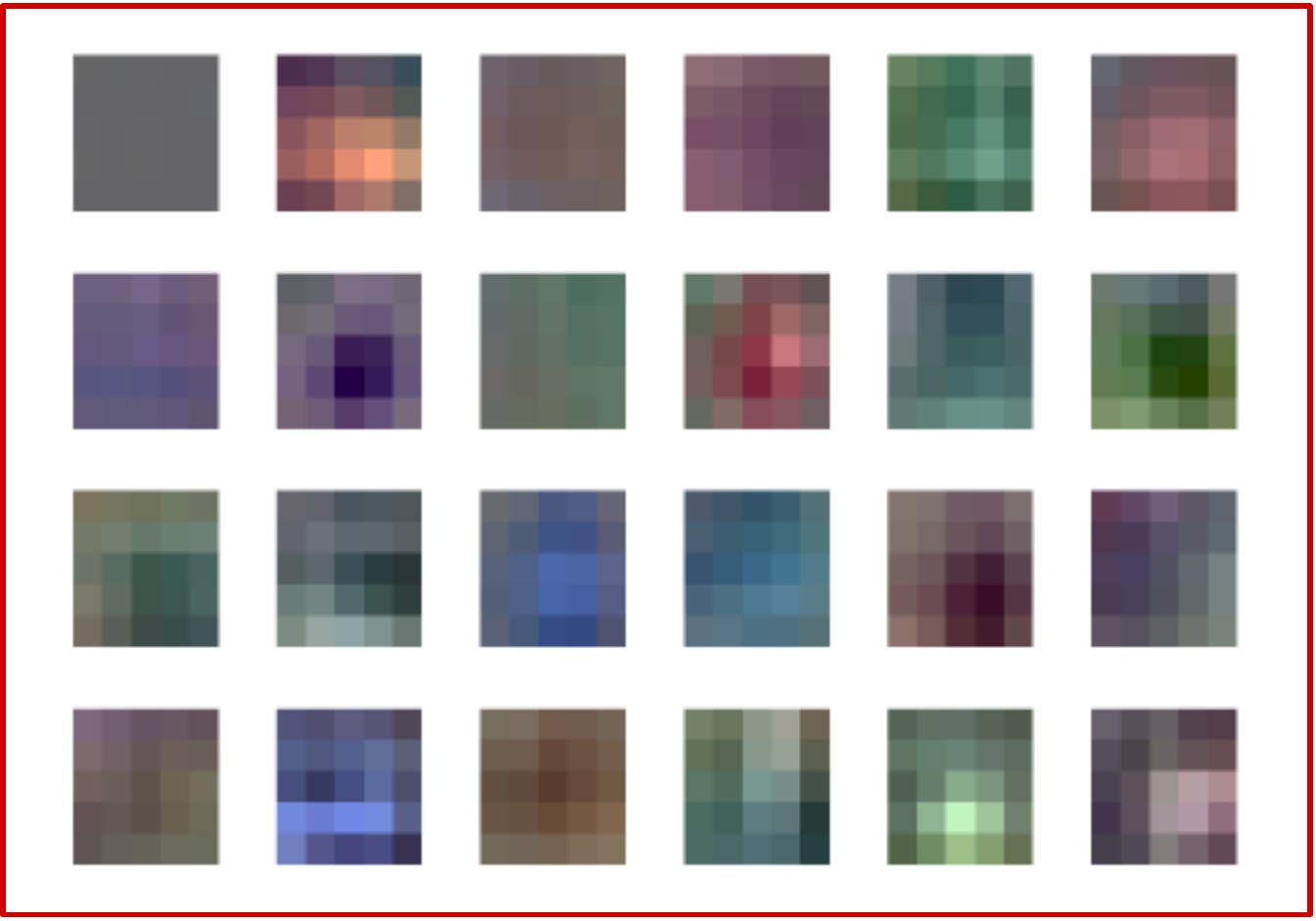}& \hfill
\includegraphics[width=0.4\textwidth]{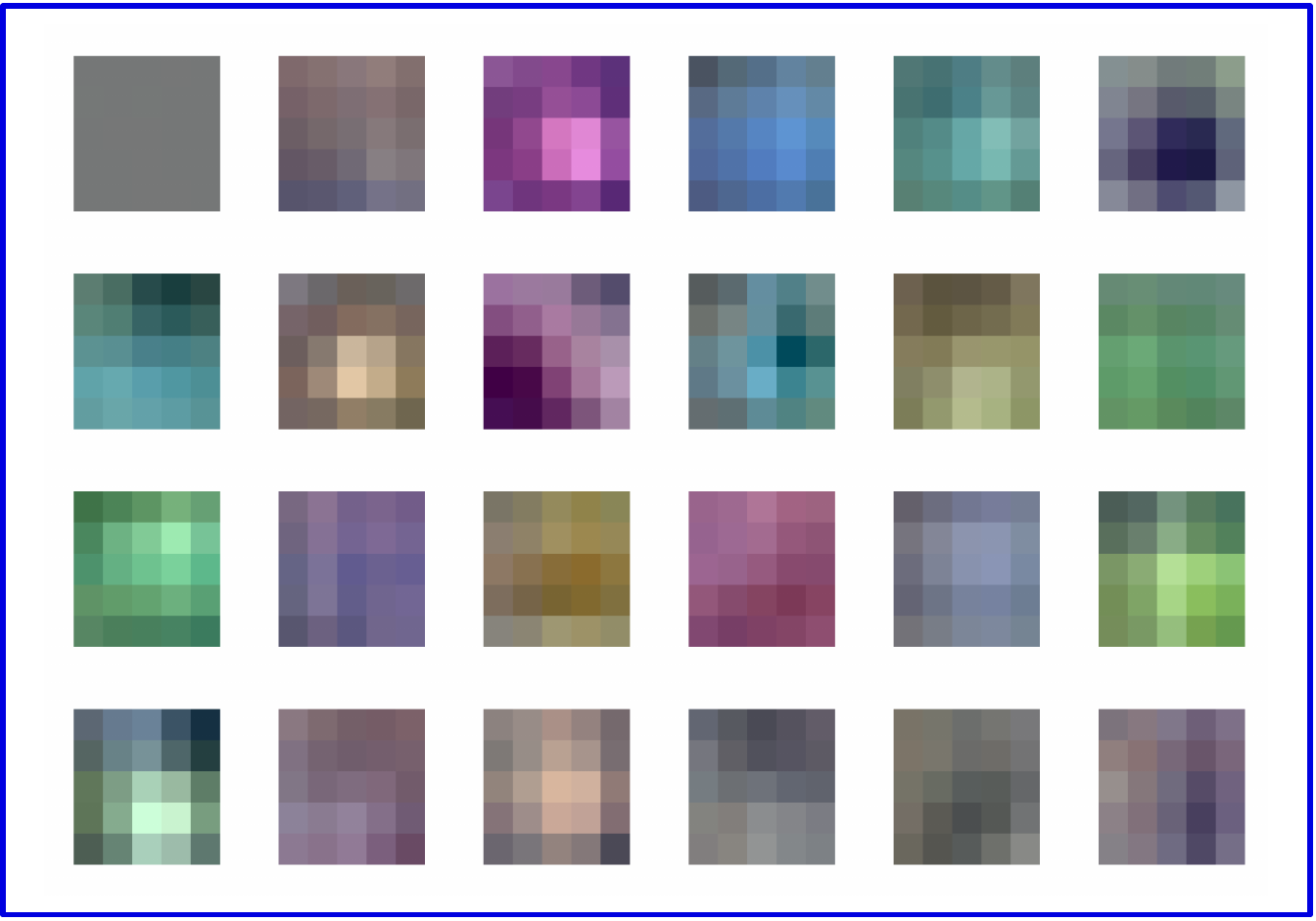}\\
(a)&(b)
\end{tabular}
\caption{Visualization of the kernels in the first convolutional layer: (a) Kernels for processing the input image. (b) Kernels that process the background image.}
\label{fig:filter_kernel_c1}
\end{figure}

For further understanding and intuition, we  visualize the convolutional filters and the feature maps when an image pair is fed into the network. In order to replace feature engineering, filters are employed which are learned during training. Since the learning is performed with ground truth data, highly application specific features will be extracted. Our network contains 3 convolutional layers which perform the feature extraction, especially the convolutional filters, the so-called kernels, are responsible for that task. Some of those will be illustrated in the following. All convolutional filters in our network are of size $5 \times 5$. The filters that are directly applied to the RGB image pair of input and background image are illustrated in Figure \ref{fig:filter_kernel_c1}.
\begin{figure}[!h]
\centering
\begin{tabular}{cc}
\includegraphics[height=120pt,width=0.45\textwidth]{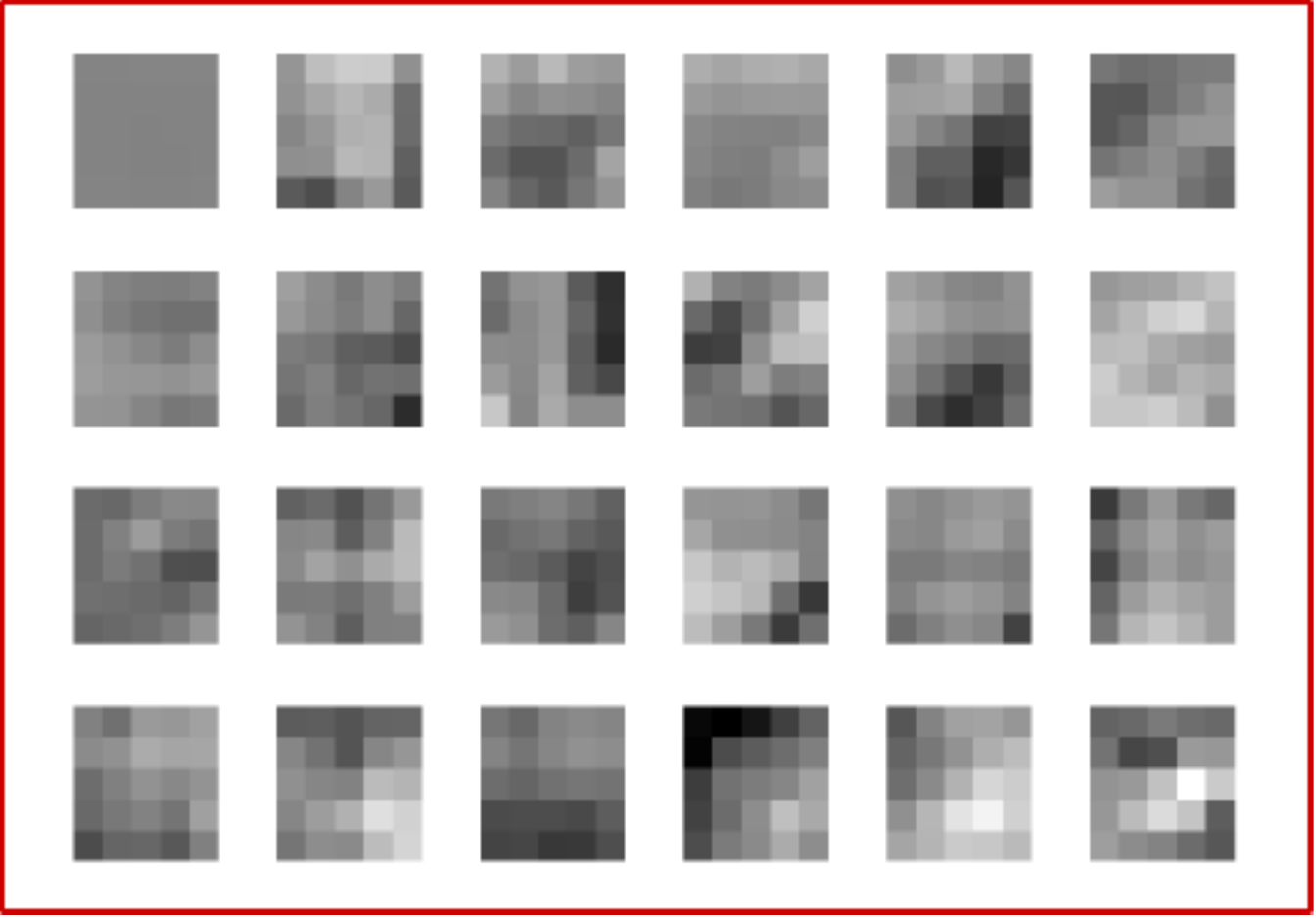}&
\includegraphics[height=240pt,width=0.45\textwidth]{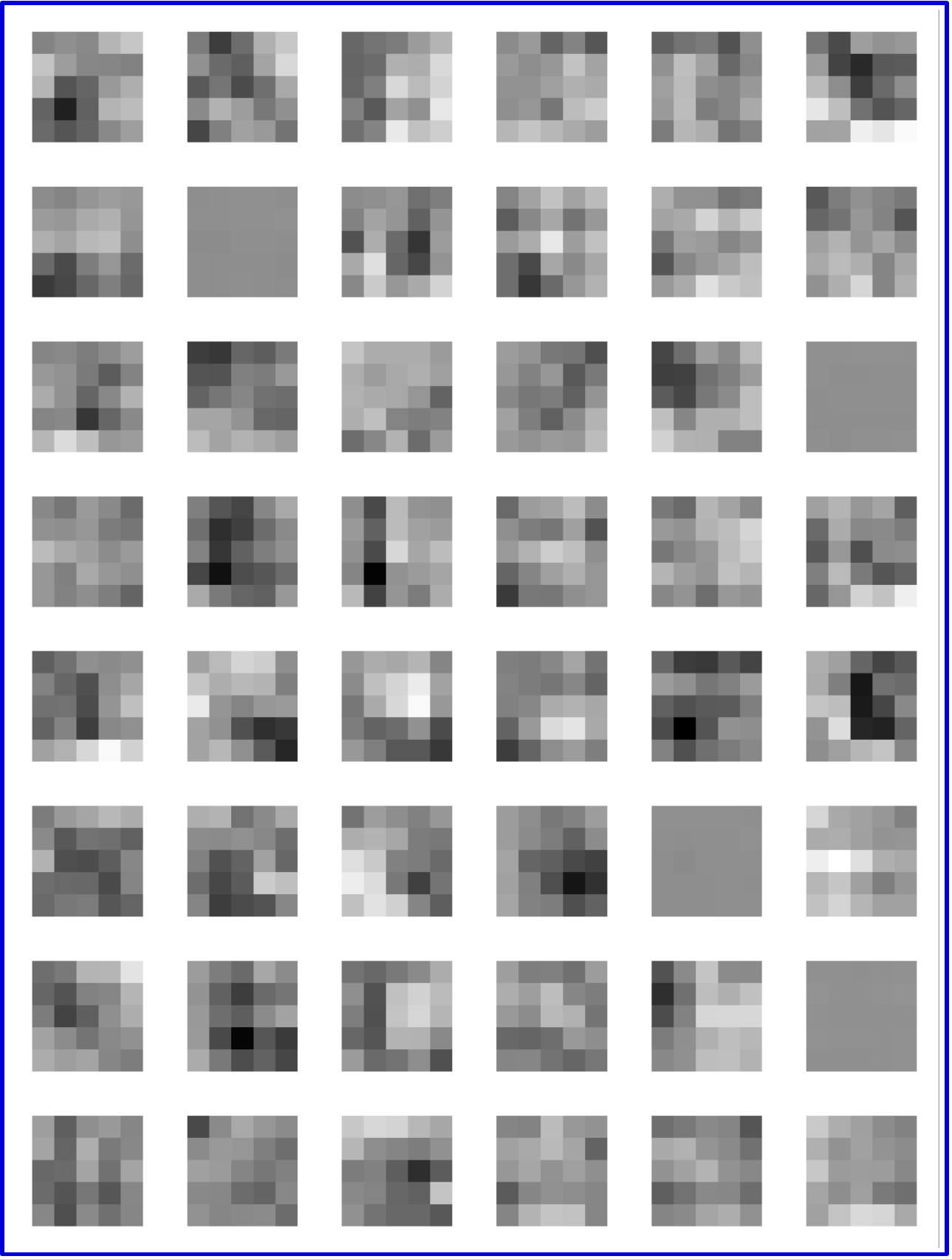}\\
(a)&(b)\\

\end{tabular}
\caption{Kernels of the second and third convolutional layer: (a) 24 kernels to generate the first output feature map in the second convolutional layer. (b) 48 kernels that output the first feature map in the third convolutional layer.}
\label{fig:filter_kernel_c2_c3}
\end{figure}\\
Due to the large number of filters, we only show the filter-sets that output the first feature map of the respective convolutional layer. Those filter-sets from the remaining convolutional layers are shown in Figure \ref{fig:filter_kernel_c2_c3}. By applying the filters onto the input images or feature maps, the convolutional layer scans the input for certain patterns, that are defined by the filters, that again generate new feature maps which capture task-specific characteristics.
\subsubsection{Feature Analysis}
By applying the learned filters to the network input, passing the intermediate values through the activation and subsampling functions, feature maps are generated. Those are again the input of the subsequent convolutional layer or classifier. In order to analyze the generated feature maps, we first need to feed an input into the network. An example input pair consisting of background and input frame is shown along with the CNN output in Figure \ref{fig:network_input_output}. The output feature maps from all convolutional layers in our network are illustrated in Figures.~\ref{fig:filter_kernel_c1} and~\ref{fig:filter_kernel_c2_c3}. At the end, the vectorized feature maps of the last convolutional layer in our CNN are fed into a MLP that performs the pixel-wise classification.


\section{Conclusion}
\label{sec:conclusion}
We have proposed a novel approach based on deep learning for background subtraction task. The proposed approach includes three processing steps, namely background model generation, CNN for feature learning and post-processing. It turned out that the CNN yielded the best performance among all other algorithms for  background subtraction. In addition, we analyzed our network during training and visualized the convolutional filters and generated feature maps. Due to the fact that our method works only with RGB data, the potential of Deep Learning approaches and the available data, one could think of modeling the background with Deep Learning techniques, for example with RNN. Furthermore, as we use a global threshold for our network outputs at the moment, one could use adaptive pixel-wise thresholds, as in the PBAS algorithm, employing a kind of ''background dynamics'' measure for the feedback loop. With this adaption, one could increase the sensitivity in static background areas and decrease it for areas with dynamic background. At last, when combining our method with an existing background subtraction algorithm, in our case with the SuBSENSE algorithm, one could use the information of both output segmentations and combine them to get a refined output and employ this output for improving the updates of the background model.

\clearpage
\bibliographystyle{abbrv}
\bibliography{DeepBS}
\end{document}